\documentclass{article}

    \usepackage[preprint]{neurips_2021}

\usepackage[utf8]{inputenc} 
\usepackage[T1]{fontenc}    
\usepackage{xr-hyper} 
\usepackage{hyperref}       
\usepackage{url}            
\usepackage{booktabs}       
\usepackage{amsfonts}       
\usepackage{nicefrac}       
\usepackage{microtype}      
\usepackage{xcolor}         

\usepackage[]{todonotes}
\usepackage{amsmath, mathtools} 
\usepackage[]{bm}
\usepackage{graphicx}
\usepackage{amssymb}
\usepackage{pifont}
\newcommand{\cmark}{\ding{51}}%
\usepackage{multirow}   
\usepackage{tikz}
\usetikzlibrary{automata, positioning, arrows} 
\usepackage{enumitem} 
\usepackage{xurl} 

\usepackage{caption} 
\usepackage{subcaption} 

\DeclareMathOperator*{\sign}{sign}
\DeclareMathOperator*{\sech}{sech}
\DeclareMathOperator*{\pcc}{pcc\;loss}
\DeclareMathOperator*{\mape}{mape\;loss}

\newcommand{\iridis}{Iridis 5 }  
\newcommand{\soton}{Southampton }  
\newcommand{\uniform}{$\mathcal{U}$}

\title{Learning Division with Neural Arithmetic Logic Modules}

\author{%
    Bhumika Mistry \& Katayoun Farrahi \& Jonathon Hare \\
    Department of Vision Learning, and Control\\
    Electronics and Computer Science\\
    University of Southampton \\
    \texttt{$\{$bm4g15, k.farrahi, jsh2$\}$@soton.ac.uk} \\
}

\begin{document}

\maketitle

\begin{abstract}
To achieve systematic generalisation, it first makes sense to master simple tasks such as arithmetic. 
Of the four fundamental arithmetic operations (+,-,$\times$,$\div$), division is considered the most difficult for both humans and computers. 
In this paper we show that robustly learning division in a systematic manner remains a challenge even at the simplest level of dividing two numbers. 
We propose two novel approaches for division which we call the Neural Reciprocal Unit (NRU) and the Neural Multiplicative Reciprocal Unit (NMRU), and present improvements for an existing division module, the Real Neural Power Unit (Real NPU). 
Experiments in learning division with input redundancy on 225 different training sets, find that our proposed modifications to the Real NPU obtains an average success of 85.3$\%$ improving over the original by 15.1$\%$. 
In light of the suggestion above, our NMRU approach can further improve the success to 91.6$\%$. 
\end{abstract}

\section{Introduction}\label{sec:introduction}
The ability to model division, one of the four fundamental arithmetic operations, is necessary for expressing dynamical systems \citep{sahoo2018-eql-div}, and physics-based formulas \citep{Udrescueaay2631}. 
\citet{madsen2020neural} states that `redundant units are very common in neural networks, which are often overparameterized’. As differentable specialist modules (e.g., for arithmetic operations) can be integrated with such overparametrized networks as an intermediate module \citet{mistry2021primer}, being able to successfully select only the relevant inputs is important. 
Selecting relevant inputs/features is a desirable property of neural networks useful for improved intepretability, reduced pre-processing costs and greater generalisation \citep{chandrashekar2014}. 
However, even recent models still struggle to learn division when there is input redundancy \citep{schlor2020inalu}. 

In particular, imagine you must learn to divide 2 numbers from a list of 10 numbers, but are only given the 10 numbers and the result of dividing the 2 values. 
This task requires finding the 2 relevant operands, the order to divide the operands, and learning to divide. 
In machine learning, this is equivalent to a supervised regression task where the aim is to learn the underlying function between the inputs and output such that the solution is generalisable to any input. 

The main challenge of the above task comes from learning the selection and operation at the same time, which can lead to conflicting priorities when learning network weights. 
Furthermore, the natural properties of division of values around zero leads to undesirable gradients. 
Models which deal with this naively (e.g. MLPs) are unable to deal with the fluctuant gradients caused by the asymptotic nature and discontinuities in division \citep{trask2018neural}. 

Can we build models which can learn division in the presence of its undesirable, yet valid, properties? We aim to address this question in this paper. 
Specifically, we contribute the following:\footnote{Code (MIT license) available at: \url{https://github.com/bmistry4/nalu-stable-exp-neurips-review}.}
\begin{itemize}[leftmargin=*]
    \item \textbf{Improvements to the Real NPU} \citep{heim2020neural} including: clipping, discretisation and constrained initialisation to improve performance in learning division on different training ranges.
    \item \textbf{Two novel division modules}, the NRU and the NMRU. The NRU explores extending the NMU weight ranges from [0,1] to [-1,1] to include division, where we find a weakness in learning from negative ranges. Learning from the weaknesses of the NRU, the NMRU extends the NMU to learn division while keeping weights values between [0,1]. We further boost performance by using a Real NPU inspired sign retrieval mechanism, enabling the NMRU to gain the best performance when using a mean squared error (MSE) loss. 
    \item \textbf{New understanding into the hindrances in learning division} including: training on mixed-sign inputs, training on negative ranges, training on different distributions, and division on extremely small values. We find these difficulties can be sufficiently identified using synthetic division tasks.
\end{itemize}

\section{Related Work}
One approach to learn division would be symbolic regression networks \citep{sahoo2018-eql-div}. 
However, a symbolic approach pre-defines the operations, which is not a limitation of using Neural Arithmetic Logic Modules (NALMs). 

NALMs are neural networks which learn arithmetic operations and input selection \citep{mistry2021primer}. 
The weights of these networks are intepretable such that a discrete value represents a specific operation. 
For example, `-1' to represent division and `0' for no selection. 
From this research field, we focus on the Real NPU and the NMU. 
Until now, the Real NPU only has learned division on training ranges of either \uniform[0.1,2] or Sobol(0,0.5) \citep{heim2020neural}. 
It remains unclear if this module is robust to other training ranges even as a stand-alone unit. 
Robustness to training ranges is important as these module's applicational use comes from being part of larger end-to-end networks, where the input range into the module cannot be controlled. 
The NMU is a multiplication module which we extend to also do division. The authors of the NMU believe such an extension incurs too many limitations for learning \citep{madsen2020neural}.
We use this paper as an opportunity to explore this belief. 

\citet{trask2018neural} developed the Neural Arithmetic Logic Unit (NALU) which can model all four arithmetic operations. However, studies show this module to be unstable in learning division \citep{schlor2020inalu, heim2020neural}. In particular, their gating method responsible for selecting an operation cannot learn consistently \citep{madsen2020neural}. 
\citet{schlor2020inalu} developed iNALU additionally applying weight and gradient clipping, sign retrieval, regularisation, reinitialisation and separating shared parameters to the NALU. 
Even with these modifications, they still find consistently learning division to a high precision to remain unattainable. 
Furthermore, \citet{heim2020neural}'s results imply iNALU is outperformed by the Real NPU for division.

\section{Architectures}\label{sec:architectures}
This section introduces the architectures for the (Real) NPU, NRU, and the NMRU.
The (Real) NPU is an existing module, which we improve in Section~\ref{sec:real-npu-mods}. 
The NRU and NMRU are novel contributions. 
Appendix~\ref{app:division-module-properties} summarises the important properties of these division modules. 

\subsection{Real Neural Power Unit}\label{sec:arch-realnpu}
\cite{heim2020neural} develop a module to learn to multiply and divide, using the intuition from \citet{trask2018neural} that \textit{multiplicative operations are additive operations in log space}. 
Their work extends this idea into complex space. 
The NPU can be used with its complex form (Equation~\ref{eq:npu}) requiring both a complex and real weight matrix ($\bm{W^{\textrm{IM}}}, \bm{W^{\textrm{RE}}}$), or only its real form the Real NPU (Equation~\ref{eq:real-npu}). 
The $\odot$ represents element-wise multiplication (Hadamard/Schur) product). 
For improved gradients, a relevance gate $\bm{r}$ (Equation~\ref{eq:npu-r}) is used which converts inputs close to 0 (i.e. irrelevant features) to 1 to avoid the resulting output evaluating to 0. 
A gating vector $\bm{g}$, learns to select relevant input elements, where gate values are clipped between [0,1] during training. 
\begin{equation}
\begin{aligned}
\mathrm{NPU} := &\exp(\bm{W^{\textrm{RE}}}\log(\bm{r})- \bm{W^{\textrm{IM}}}\bm{k}) \, \odot \cos(\bm{W^{\textrm{IM}}}\log(\bm{r}) + \bm{W^{\textrm{RE}}k}) , \label{eq:npu} 
\end{aligned}
\end{equation}
\begin{equation}
\begin{aligned}
\mathrm{Real\; NPU} := &\exp(\bm{W^{\textrm{RE}}}\log(\bm{r})) \, \odot \cos(\bm{W^{\textrm{RE}}k}) 
\label{eq:real-npu}
\end{aligned}
\end{equation}
\begin{align}
\textrm{where} \quad
\bm{r} &= \bm{g} \odot (|\bm{x}|+ \epsilon) + (\bm{1} - \bm{g}) \label{eq:npu-r} 
\quad \textrm{and}  \quad
k_i  = 
    \begin{cases}
       0 & x_i \geq 0 \\
       \pi\mathrm{g_i} & x_i < 0
    \end{cases} \;.
\end{align}
A weighted L1 penalty is used when training. 
The weight value $\beta$ grows between predefined values $\beta_{start}$ to $\beta_{end}$ and is increased every $\beta_{step} = 10,000$ iterations  by a growth factor $\beta_{growth} = 10$. 
We focus on the Real NPU over the NPU as the solution of the tasks in this paper can be captured using only real values meaning that the complex form is not required. 

\subsection{Neural Reciprocal Unit}\label{subsec:arch-nru}
We propose the NRU, which can model multiplication and division.  
We extend the NMU, motivated by \textit{division being multiplication of reciprocals}. 
The range which weight values can be is extended from [0,1] to [-1,1], where -1 represents applying the reciprocal on the corresponding input element. 
A NRU output element $z_o$ is defined as 
\begin{align}
\textrm{NRU}: z_o &= \prod_{i=1}^{I} \left(\sign(\mathrm{x}_{i})\cdot|\mathrm{x}_{i}|^{W_{i,o}}\cdot|W_{i,o}| + 1 - |W_{i,o}|\right) , \label{eq:hard-nru}
\end{align}
where $I$ is the number of inputs. 
Assuming weights are either 1 (multiply) or -1 (reciprocal), $|\mathrm{x}_{i}|^{W_{i,o}}$ will apply the operation on an input element. 
The absolute value is used so that the module only operates in the space of real numbers, as $x_i^{W_{i,o}}$ for a negative input ($x_i$) when $-1 < W_{i,o} < 1$ results in a complex number. 
The use of absolute means the sign of the input must be reapplied. 
For the no-selection case $W_{i,o}=0$, we want the input element to convert to 1 (the identity value), resulting in applying $\cdot|W_{i,o}| + 1 - |W_{i,o}|$. 
The derivative of the absolute function at 0 is undefined meaning the gradients of Equation~\ref{eq:hard-nru} can contain points of discontinuity. 
To alleviate this issue, we approximate the absolute function using a scaled $\tanh$ (inspired by \citet{faberMarch2020nsr}). 
More formally, 
\[
    |W_{i,o}|= 
\begin{cases}
    \tanh(1000\cdot W_{i,o})^2      &  \text{if training}\\
    |W_{i,o}|                       & \text{otherwise}
\end{cases}.
\]
The scale factor (1000) controls how close to the absolute function the approximation is, where larger values give a more accurate approximation. 
For clipping and regularisation, the same scheme as the Neural Addition Unit (NAU) (see Appendix~\ref{app:nau-nmu}) is used. 

\subsection{Neural Multiplicative Reciprocal Unit}\label{subsec:arch-nmru}
An alternate extension of the NMU, also motivated by \textit{division being multiplication of reciprocals} is the NMRU (Equation~\ref{eq:nmru}). 
We concatenate the reciprocal of the input (plus a small $\epsilon$) to the input resulting in a module which only needs to learn selection. Hence, weights can be in the range [0,1]. 
\begin{align}
\textrm{NMRU}: z_o &= \prod_{i=1}^{2I} \left(W_{i,o}\cdot |\mathrm{x}_{i}| + 1 - W_{i,o} \right) \cdot \cos(\sum_{i=1}^{2I}\left(W_{i,o}\cdot k_i)\right)\;, \label{eq:nmru}
\textrm{where }
k_i & = 
    \begin{cases}
       0    & x_i \geq 0 \\
       \pi  & x_i < 0
    \end{cases} \;.
\end{align} 
The iteration over $2I$ represents the going through all inputs and their reciprocals. 
We calculate the magnitude and sign separately, joining the result at the end. 
The magnitude is calculated passing absolute of the concatenated input through an NMU architecture and the sign by using a cosine mechanism similar to the Real NPU. 
However, unlike the Real NPU only the weight matrix is required. 
The norm of the weight's gradients are clipped to 1 prior to being updated by the optimiser. 
This is done to alleviate the issue of exploding gradients caused by including the reciprocal to the inputs. 
For  clipping and regularisation, the same scheme as the NMU (see Appendix~\ref{app:nau-nmu}) is used. 

\section{Experiment Setup}\label{sec:exp-setup}
We introduce the two main experiments used to evaluate modules, including: default parameters, train and test ranges, and evaluation metrics. 
The tasks evaluate the ability of a single module to divide two numbers from an input vector in two settings: \textbf{no redundancy} and \textbf{with redundancy}.  

\paragraph{Default parameters:}
All experiments use a mean squared error (MSE) loss with an Adam optimiser \citep{Kingma2015AdamAM}, with 10,000 samples for the validation and test sets. The best model for evaluation is taken using early stopping on the validation set. 
All runs are over 25 different seeds.
All inputs are required in the \textit{no redundancy} setting, i.e., input size of 2. 
Training takes 50,000 iterations where each iteration consists of a different batch of size 128. 
The Real NPU uses a learning rate of 5e-3 with sparsity regularisation scaling during iterations 40,000 to 50,000.
The NRU and NMRU use sparsity regularisation scaling during iterations 20,000 to 35,000 and a learning rate of 1 and 1e-2 respectively. 
In contrast, the \textit{redundancy} setting uses an input size of 10, where 8 input values are not required for the final output. 
The total training iterations are extended to 100,000 with batch sizes of 128. 
The learning rates for the Real NPU, NRU and NMRU are 5e-3, 1e-3 and 1e-2 respectively. 
Sparsity regularisation scaling occurs during iteration 50,000 to 75,000 for all modules. 
A summary of all relevant parameters is found in Appendix~\ref{app:parameters}. 

\paragraph{Ranges:}
The interpolation (train/validation) and extrapolation (test) ranges, are found in Table~\ref{tab:SLTR-ranges}. The chosen ranges are influenced by \citet{madsen2020neural}.
\begin{table}[]
\centering
\caption{Interpolation (train/validation) and extrapolation (test) ranges used. Data (as floats) is drawn from a Uniform distribution with the range values as the lower and upper bounds.}
\vspace{1em}
\label{tab:SLTR-ranges}
\begin{tabular}{llllll}
\toprule
\textbf{Interpolation} & {[}-20, -10) & {[}-2, -1) & {[}-1.2, -1.1) & {[}-0.2, -0.1) & {[}-2, 2)                  \\
\textbf{Extrapolation} & {[}-40, -20) & {[}-6, -2) & {[}-6.1, -1.2) & {[}-2, -0.2)   & {[}{[}-6, -2), {[}2, 6){]} \\\midrule
\textbf{Interpolation} & {[}0.1, 0.2) & {[}1, 2) & {[}1.1, 1.2) & {[}10, 20) &  \\
\textbf{Extrapolation} & {[}0.2, 2)   & {[}2, 6) & {[}1.2, 6)   & {[}20, 40) & 
\\\bottomrule
\end{tabular}
\end{table}

\paragraph{Evaluation metrics:}
We use the \citet{maep-madsen-johansen-2019}'s evaluation scheme, consisting of three evaluation metrics: the success on the extrapolation dataset against a near optimal solution (\textit{success rate}), the first iteration which the task is considered solved (\textit{speed of convergence}), and the extent of discretisation towards the weights' inductive biases (\textit{sparsity error}). 
Sparsity error calculated by $\max\limits_{i,o}(\min(|W_{i,o}|, 1-|W_{i,o}|))$, measures the weight element which is the furthest away from the acceptable discrete weights for the module. 
A success means the MSE of the trained model is lower than a threshold value (i.e. the MSE of a near optimal solution). 
We differ from \citet{maep-madsen-johansen-2019} by using a fixed threshold value 1e-5 rather than a simulated MSE, as there are no intermediate layers to accumulate numerical errors. 
We choose this precision as it can be guaranteed when working with 32-bit PyTorch Tensors. 
95\% confidence intervals (over the 25 seeds) are calculated from a specific family of distributions dependant on the metric. 
The success rate uses Binomial distribution because trials (i.e. run on a single seed) are either pass/ fail situations.
The convergence metric uses a Gamma distribution and sparsity error uses a Beta distribution. 
Both Beta and Gamma can easily approximate the normal distribution and support its corresponding metric.

\section{Improving the Real NPU's Robustness}\label{sec:real-npu-mods}
We first improve the robustness of the Real NPU on different training ranges. 
We use the Single Module Task with no redundancy (see Section~\ref{sec:exp-setup}) to investigate the following questions: 
\begin{enumerate}
    \item Is L1 regularisation required, and if so, do the regularisation parameters require tuning?
    \item Does clipping the weight matrix aid learning? 
    \item Does enforcing discretisation on parameters improve convergence?
    \item Can the weight matrix initialisation be improved?
\end{enumerate}

To address each question in order, we propose applying incremental modifications to the Real NPU.
These modifications include: ablation study on the L1 regularisation (including a sweep over the scaling range hyperparameters), clipping, enforcing discretisation, and a more restrictive initialisation scheme. 
We assume that we are optimising the Real NPU to perform multiplication or division.
Therefore, \textit{we trade-off the flexibility of having non-discretised weights, which enables the success of modelling the SIR data in \citet[Section~4.1]{heim2020neural} , in favour of sparse models with discrete weight values.}
All the modifications suggested can also be generalised for the NPU architecture.

\paragraph{Is L1 regularisation required? (Yes)}
\begin{figure}
\centering
\subfloat[L1 regularisation]{\includegraphics[width=0.31\linewidth]{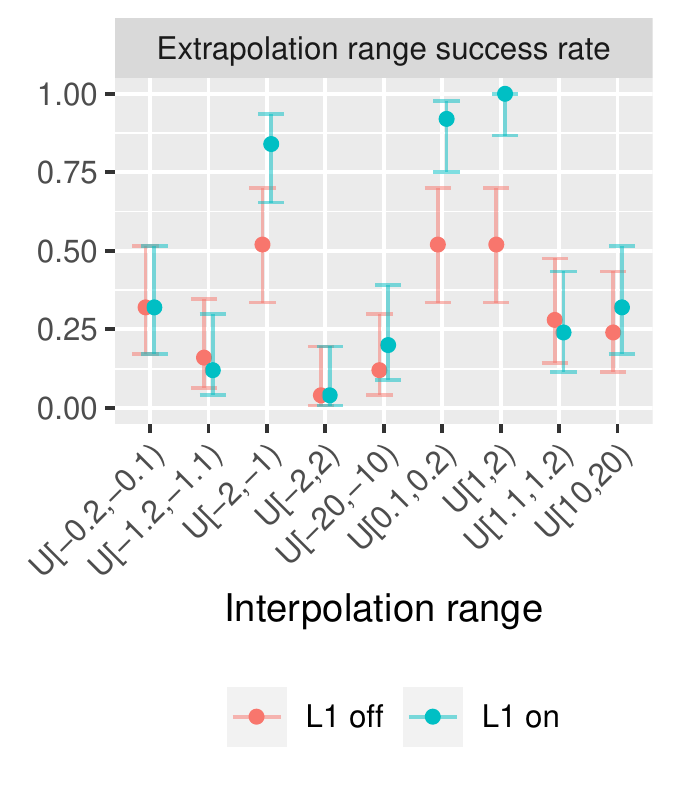}\label{subfig:realnpu-L1T-vsL1F}}\hspace{0.2cm}
\subfloat[Sweep over L1 (start,end) beta parameters]{\includegraphics[width=0.44\linewidth]{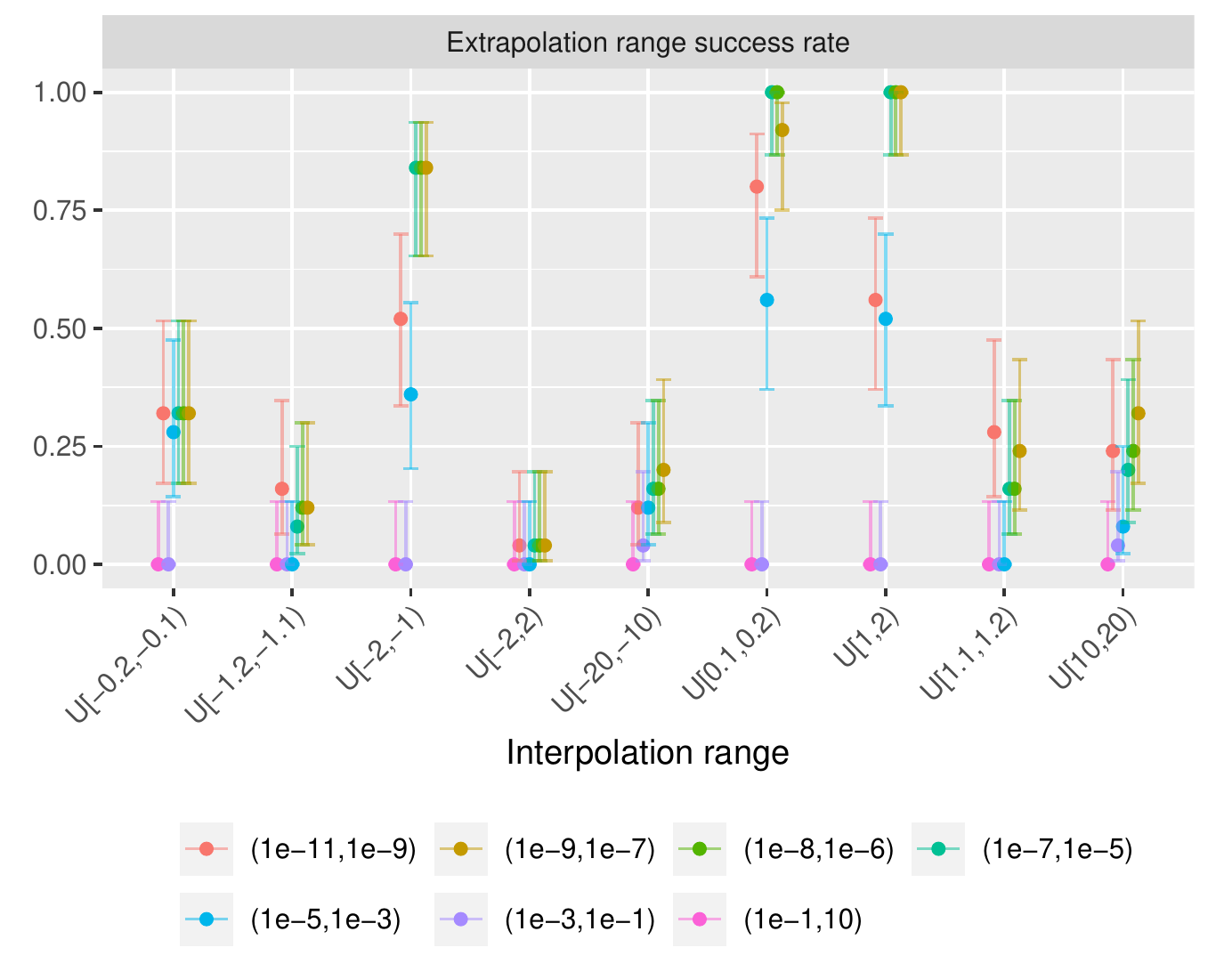}\label{subfig:realnpu-L1-sweep}} 
\caption{Exploring the effect and sensitivity of L1 regularisation on the Real NPU}\label{fig:realnpu-L1}
\end{figure}
L1 encourages sparsity (i.e., zero weights) in solutions. 
Zero-valued weights means not to select an input and return the identity value 1. 
For the task, the optimal weight values require selecting all inputs and therefore non-zero values, suggesting the application of L1 could be damaging. 
Therefore, we compare against a model which does not use L1 regularisation, shown in Figure~\ref{subfig:realnpu-L1T-vsL1F}. 
Removing L1 proves to be detrimental in five of the nine cases shown and only shows minor improvements in two of the nine ranges (i.e., \uniform[-1.2,-1.1) and \uniform[1.1,1.2)). 
Hence, we keep L1 regularisation.\footnote{We also experimented with using L2 regularisation but found L1 to perform better. Results are found in Appendix~\ref{app:realnpu-in2-extra}.} 
The L1 regularisation scaling (see Section~\ref{sec:arch-realnpu}), requires setting the hyperparameters for the start ($\beta_{start}$) and end ($\beta_{end}$) scaling values. 
We run a sweep over six different start and end values, denoted (<start>, <end>), displaying results in Figure~\ref{subfig:realnpu-L1-sweep}. 
We find the configuration (1e-9, 1e-7) is the most successful when considering performance on all the ranges, and larger scaling values perform worse. 

\paragraph{Does clipping the learnable parameters help? (Yes)}
\begin{figure}
\centering
\subfloat[Clipping]{\includegraphics[width=4.5cm]{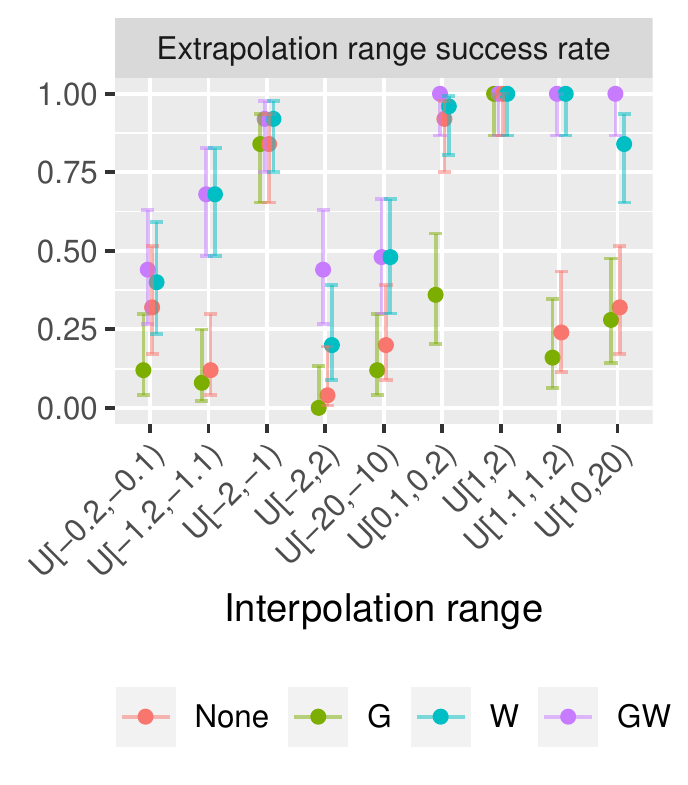}\label{subfig:realnpu-clipping}} \hspace{0.2cm}%
\subfloat[Discretisation regularisation]{\includegraphics[width=4.5cm]{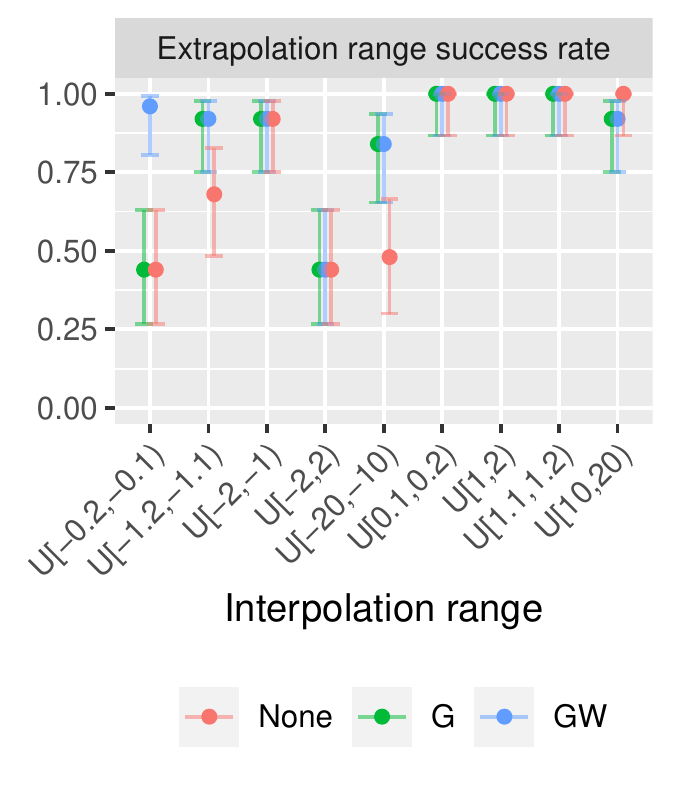}\label{subfig:realnpu-discretisation-reg}}   \hspace{0.2cm}%
\subfloat[$W^{\textrm{RE}}$ initialisation schemes]{\includegraphics[width=4.5cm]{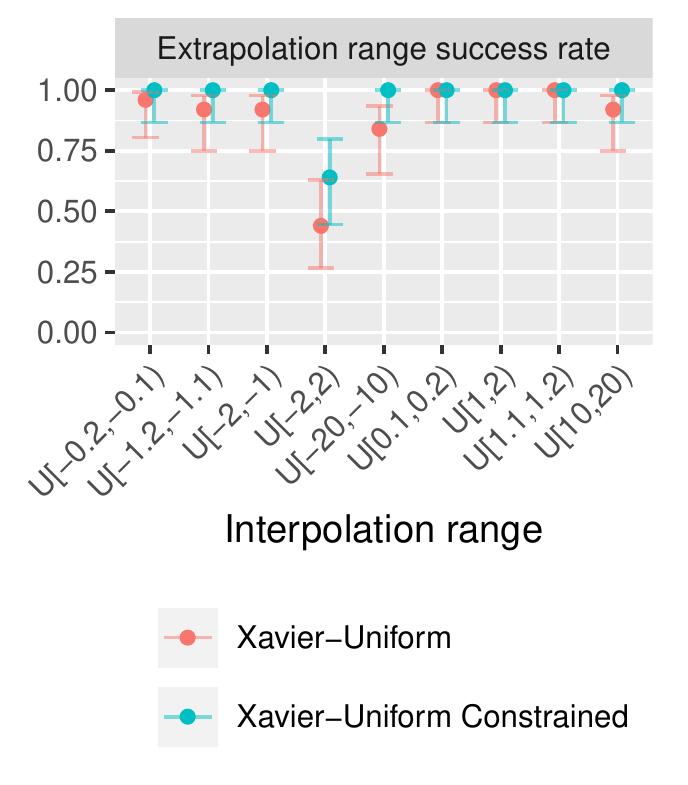}\label{subfig:realnpu-init}}
\caption{Effect of clipping, discretisation, and the NAU initialisation scheme on the Real NPU.}\label{fig:realnpu-modifications}
\end{figure}
Division and multiplication operations are represented by weight values of -1 and 1 respectively. 
The current architecture does not constrain the weights which can result in large weight values. 
The gate weights do get clipped and saved to another variable during the forward pass, meaning after an update step the gate values can also be out of the range [-1,1]. 
Hence, we investigate the effect of applying clipping directly to the weight and gate values after every optimisation step. 
Results, shown in Figure~\ref{subfig:realnpu-clipping}, show clipping is beneficial, with clipping on both weight and gate (or just on the weights) to improve over the baseline on all ranges (excluding \uniform[1,2) where the baseline has already achieved full success). 

\paragraph{Does enforcing discretisation help? (Yes)}
Modelling division in a generalisable manner requires all learnable parameters to be discrete i.e., a value from \{-1, 0, 1\}.
Using \citet{madsen2020neural}'s regularisation scaling scheme, we penalise weights for not being discrete. 
We modify the scaling factor to be $\hat{\lambda}=1$ and the regularisation to go from `off' to `on' between iterations 40,000 to 50,000. 
Results, shown in Figure~\ref{subfig:realnpu-discretisation-reg}, show discretising the gate improves over the baseline but also discretising the weights is additionally beneficial (especially for range \uniform[-0.2,-0.1)). 
\uniform[10,20) is the only range where the baseline outperforms using discretisation, succeeding on two additional seeds. 

\paragraph{Does using a more constrained initialisation help? (Yes)}
$\bm{W^{\textrm{RE}}}$ uses a Xavier-Uniform initialisation~\citep{glorot2010understanding}. 
This can result in weights initialised out of the range [-1,1]. 
Therefore, we use the initialisation for the Neural Addition Unit which is a constrained form of the Xavier-Uniform that does not allow the fan values of the uniform distribution to go beyond 0.5, meaning that no weight value will be out of the range [-1,1]~\citep{madsen2020neural}. 
Figure~\ref{subfig:realnpu-init} shows using the constrained initialisation provides improvements over multiple ranges.

\section{Results: Single Module Task}\label{sec:results}
We analyse the results for the: Real NPU without using the modifications of Section~\ref{sec:real-npu-mods}, Real NPU with modifications, NRU, and NMRU. 
\subsection{No Redundancy}
\begin{figure}
  \centering
  \includegraphics[width=\textwidth]{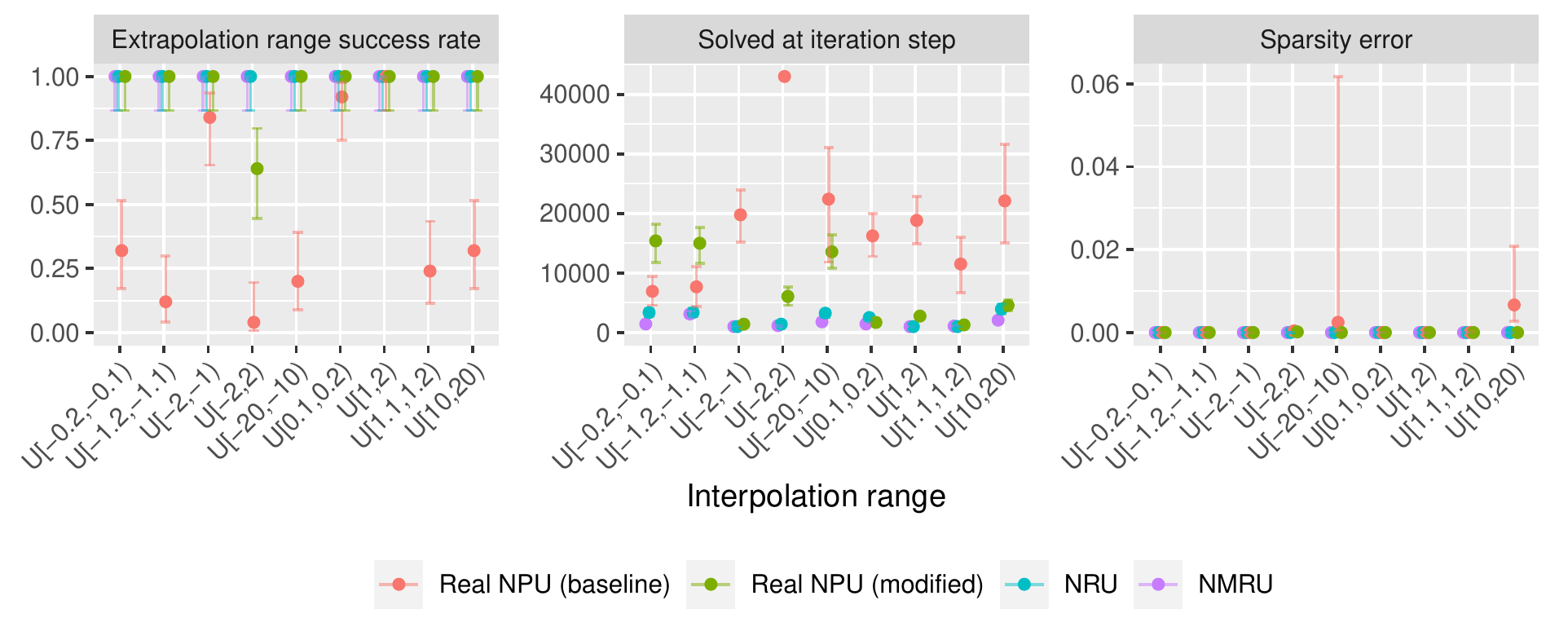}
  \caption{Division without redundancy (input size 2).}
  \label{fig:sltr-in2}
\end{figure}
Figure~\ref{fig:sltr-in2} shows the baseline Real NPU without modifications struggles with all ranges except \uniform[1,2), struggling with sparsity on the larger ranges. 
Applying the modifications deals with the sparsity issue and improves the robustness such that only range \uniform[-2,2) struggles (with a success rate of 0.64). 
The NRU and NMRU achieve full success over all ranges while solving the problem consistently fast and with low sparsity error. 
The success of the NRU is correlated with the learning rate (see Appendix~\ref{app:nru-in2-lr}).
\subsubsection{Mixed-signed Inputs} 
\begin{figure}
  \centering
  \includegraphics[width=\textwidth]{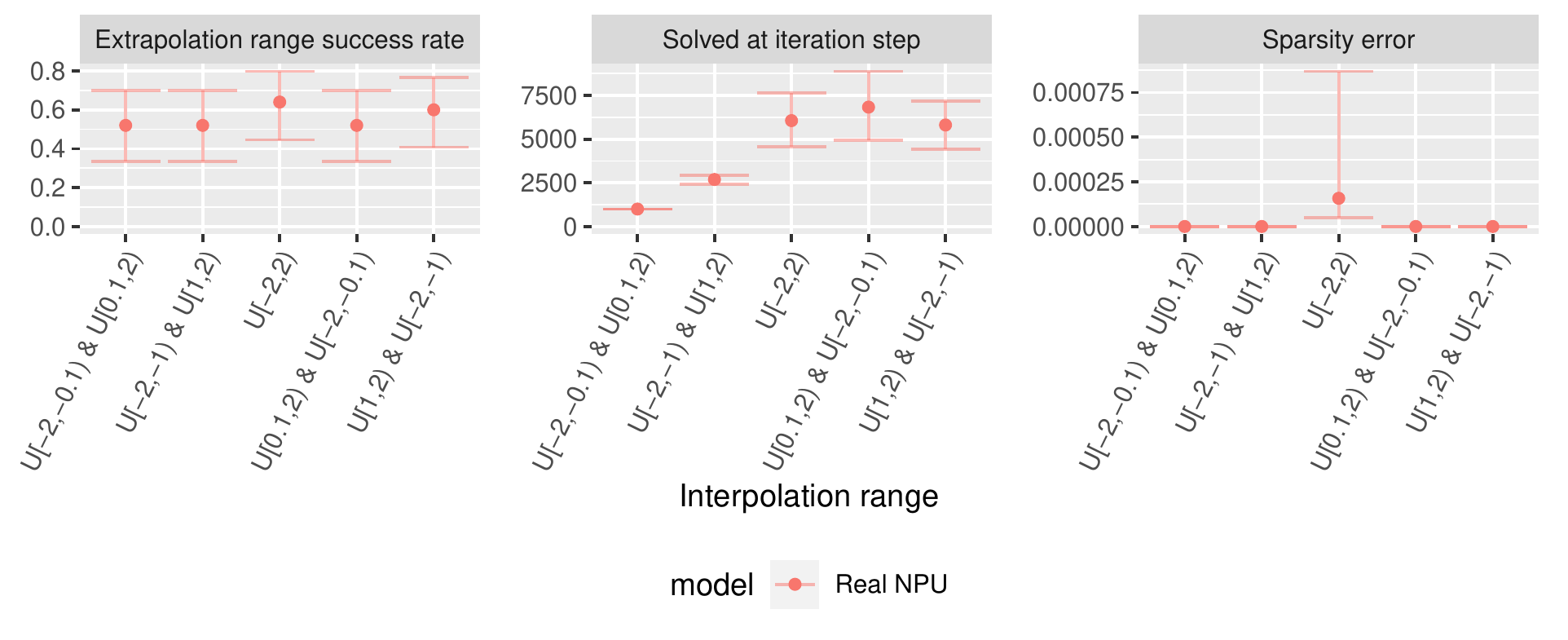}
  \caption{Extrapolation results on training the Real NPU using mixed-sign datasets that control the sign of the input elements. The ranges are in order of the datasets (i.e. dataset 1 to 5).}
  \label{fig:mixed-signs}
\end{figure}

The remaining failure range of the Real NPU is \uniform[-2,2) where inputs can consist of arbitrary signed values (e.g. all positives, all negatives, or a mixture of positive and negative values).
\textit{We question if the failure is due to the input samples in a batch having different signs from each other, or if the problem is due to the fact data samples can be close to 0 (leading to singularity issues).} 
To investigate this, we create additional mixed-sign datasets, controlling the range for each element in the input. 
The interpolation and extrapolation ranges for the different datasets can be found in Appendix~\ref{app:parameters}. 
Datasets 1, 2, 4 and 5 sample a positive value for one input element and a negative value for the other element. 
Dataset 3 samples the signs randomly. Datasets 2 and 5 avoid sampling close to 0 values to mitigate the singularity issue.  
As shown by Figure~\ref{fig:mixed-signs}, the Real NPU struggles on all these ranges, implying that the core issue is not from different input samples having different signs or due to the input samples being able to contain small values close to 0. 
The underlying issue is therefore most likely correlated to the each element in an input having different signs. 
When the denominator of the output is positive (dataset 1 or 2), the solution is found faster than when the denominator is a negative value (dataset 4 or 5). 
When the signs for an input element are controlled, discretisation/sparsity is no problem, in contrast when the signs are arbitrary the sparsity error are slightly (though not significantly) higher.

\subsection{Division by Small Numbers}
Division by zero remains a challenge to model due to the inability to provide an computational value for the output and gradient. Furthermore, the discontinuous nature at zero causes its neighbouring values to have large gradients. 
To understand the extent of this issue when learning, we explore learning to divide by values close to zero using three tasks with increasing difficulty: 1) learning to take the reciprocal of a single input, 2) taking the reciprocal of the first input given two inputs, and 3) diving the first input by the second given two inputs. 
\begin{figure}
  \centering 
  \includegraphics[width=\textwidth]{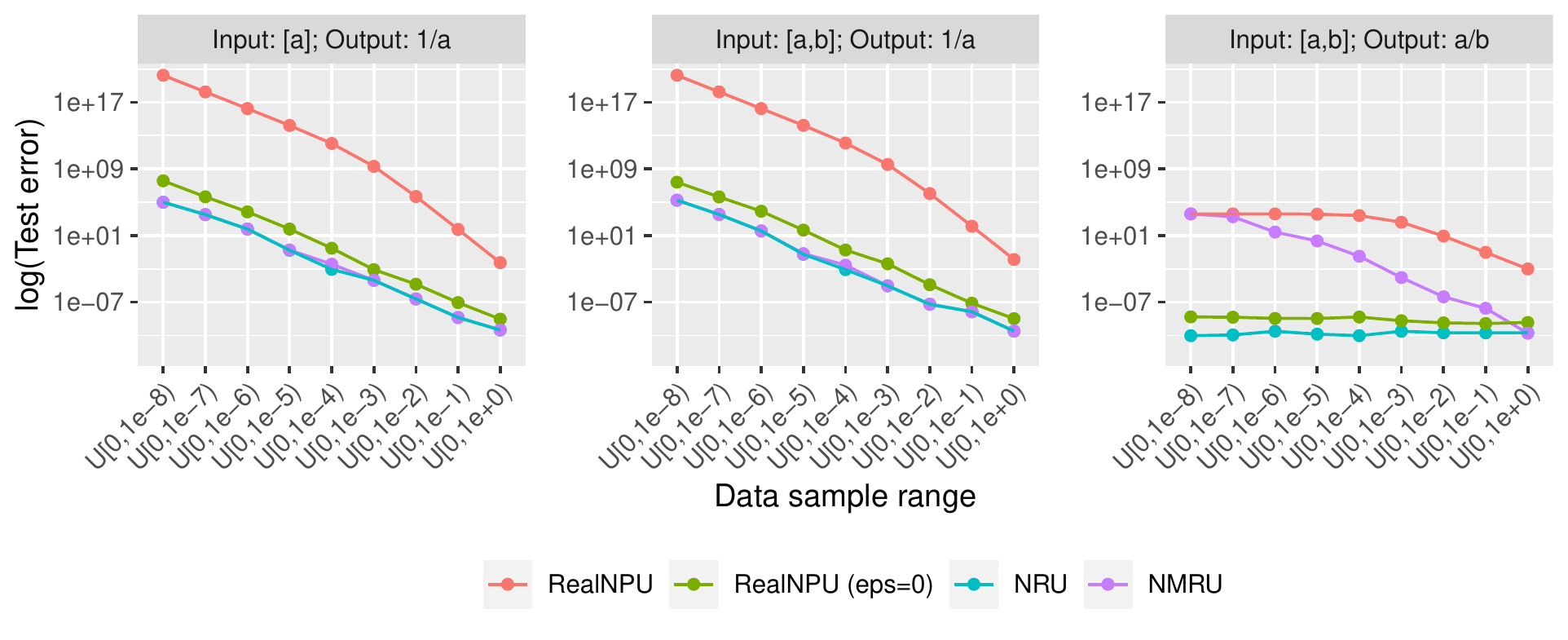}
  \caption{Effect of the singularity issue on the Real NPU, NRU and NMRU over increasing input ranges. Left: Reciprocal for an input size of 1 (no redundancy). Middle: Reciprocal for an input size of 2 (with redundancy). Right: Division for an input size of 2 (no redundancy).}
  \label{fig:divBy0-gold-errs}
\end{figure}
Figure~\ref{fig:divBy0-gold-errs} plots the test error for different modules assuming the module weights are set to the `gold' solution for the three tasks. 
As the range values become closer to zero, the test error thresholds become increasingly large. 
Therefore, even with the correct weights, relying on the test errors alone as an indicator become increasingly deceptive with values close to zero.
The Real NPU has larger test errors for all tasks and ranges, caused by adding $\epsilon$ to the input (see Equation~\ref{eq:npu-r}). 
Setting $\epsilon=0$ reduces the test error at the cost of the ability to deal with zero-valued inputs. 
Appendix~\ref{app:divBy0} provides the corresponding experimental results for these tasks. 

\subsection{With Redundancy}
\begin{figure}
  \centering
  \includegraphics[width=\textwidth]{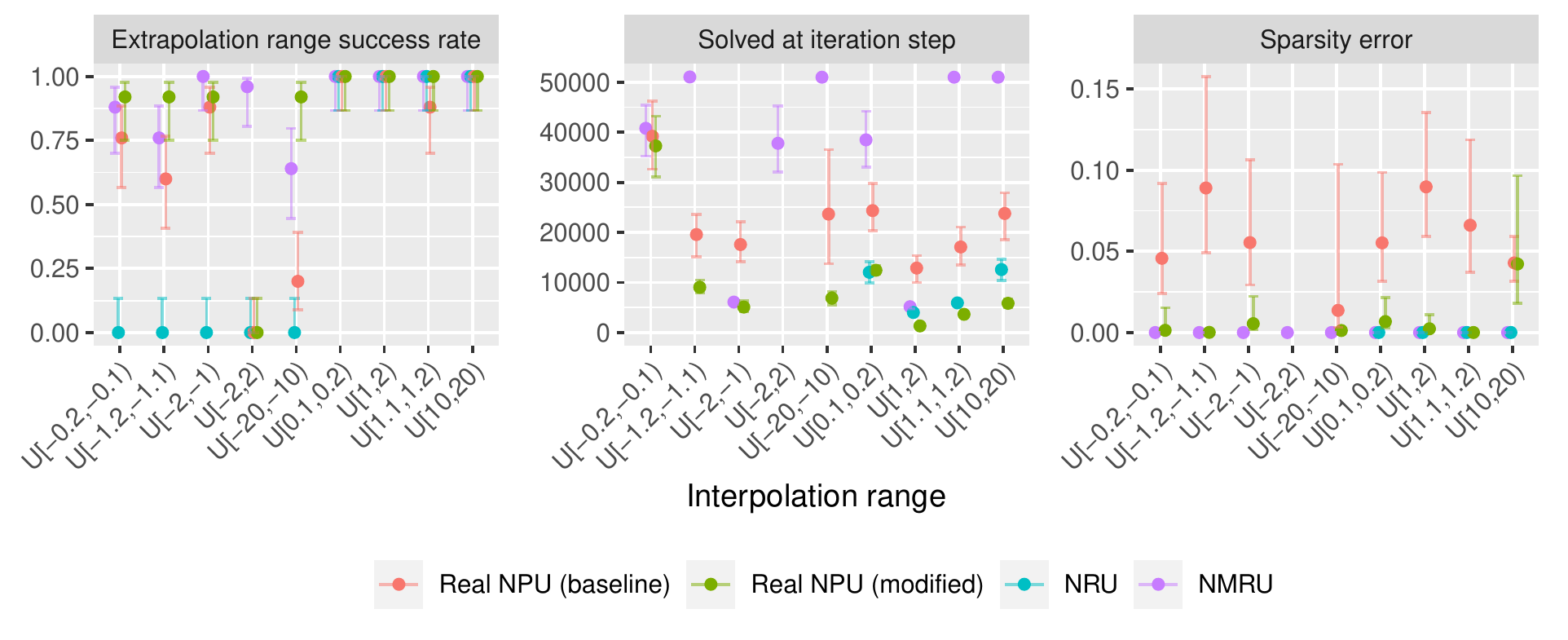}
  \caption{Division with redundancy (input size 10).}
  \label{fig:sltr-in10}
\end{figure}
Introducing redundancy (Figure~\ref{fig:sltr-in10}) causes failure modes to arise. 
Failures on range \uniform[-2,2) become more prevalent. 
The baseline Real NPU produces high sparsity errors relative to the other modules suggesting struggle with discretisation. 
Using the modified Real NPU improves over all ranges of the baseline (which were not already at full success) in terms of success, speed and sparsity.\footnote{Except for the sparsity error for range \uniform[10,20).}  
To ensure that complex weights do not fix the issue, we test the NPU module with all the modifications used on the real weight matrix (see Appendix~\ref{app:realnpu-in10-extra}). 
Complex weights hinders success and convergence speeds of negative ranges. 
Assuming the global solution only uses the real weights, we enforce the complex weights to be clipped between [-1,1] and to go to 0 during the regularisation stage using a L1 penalty. This did not result in any significant improvements against the Real NPU results. 
Input redundancy effects the NRU the most, resulting in full failures on all the negative ranges. 
The NMRU is the only module with success for the range \uniform[-2,2), which is a result of using the sign mechanism (see Appendix~\ref{app:nmru-extra-exps}). 
It performs well over all ranges though can be outperformed by the modified Real NPU for negative ranges. 
Multiple ranges for the NMRU are solved around 50,000 iterations correlating to the sparsity regularisation being turned on. 

\subsubsection{Gradient Difficulties with the NRU}
The partial derivative for the NRU weights, Equation~\ref{eq:nru-wi-grad}, can give insight to the struggles of the NRU.
\begin{equation}
\begin{aligned}
\frac{\partial \bf{\hat{y}}}{\partial w_i} = \tanh(1000w_i)(\sign(x_i)|x_i|(\tanh(1000w_i)\log(|x|) + \\ 2000\sech(1000w_i)^2) - 2000\sech(1000w_i)^2) \times \textcolor{red}{\textrm{NRU}_{\bf{\Tilde{x}} \in \bf{x}\backslash\{x_i\} }(\bf{\Tilde{x}})} .
\label{eq:nru-wi-grad}
\end{aligned}
\end{equation}

$\textcolor{red}{\textrm{NRU}_{\bf{\Tilde{x}} \in \bf{x}\backslash\{x_i\} }(\bf{\Tilde{x}})}$ applies the NRU to all inputs excluding $x_i$ influencing the gradient values between subsequent update steps. 
Factoring out this term, the following observations are made.
If $x_i\approx0$ and $w_i\approx0$ then gradients become increasingly large. 
If $x_i\approx0$ and $-1\leq w_i < 0$ then as $w_i \mathrel{\rightarrow} -1$ all gradients for $x_i$ where $|x_i|>>1$ become increasingly small.  
The gradients for $x_i=-1$ and $x_i=1$ are 0 regardless the value of $w_i$. 
If $w_i=0$ then the gradient is 0 for all $x_i$, a result of using the $\tanh$ approximation.  
Even if the sign and magnitude are calculated separately and then combined (see Appendix~\ref{app:nru-sepSigns}) to try to control the gradient better, the problem remains. 
Therefore, we conclude that extending the NMU to divide using a weight of -1 is a poor choice when there are redundant inputs.

\subsubsection{The Real NPU's and NMRU's Exploitation of Multiplicative Rules}
The NMRU solutions exploit the inverse rule of division in that $a_i \cdot \frac{1}{a_i}=1$. Since the input also contains the reciprocals, numerous extrapolative solutions exist. 
However this comes at the cost of finding a `simple' solution which contains ones only for relevant inputs.
The Real NPU exploits the rules $a_i \cdot 0=0$ and $1^{a_i}=1$ enabling non-zero weight values if the corresponding gate value is 0. 
However, we can avoid this by allowing 0 to also not be penalised during sparsity regularisation stage (see Appendix~\ref{app:realnpu-in10-extra}). We find this alleviates the exploitation issue with no cost to performance.

\section{More Challenging Distributions}
\begin{figure}[b]
  \centering
  \includegraphics[width=0.9\textwidth]{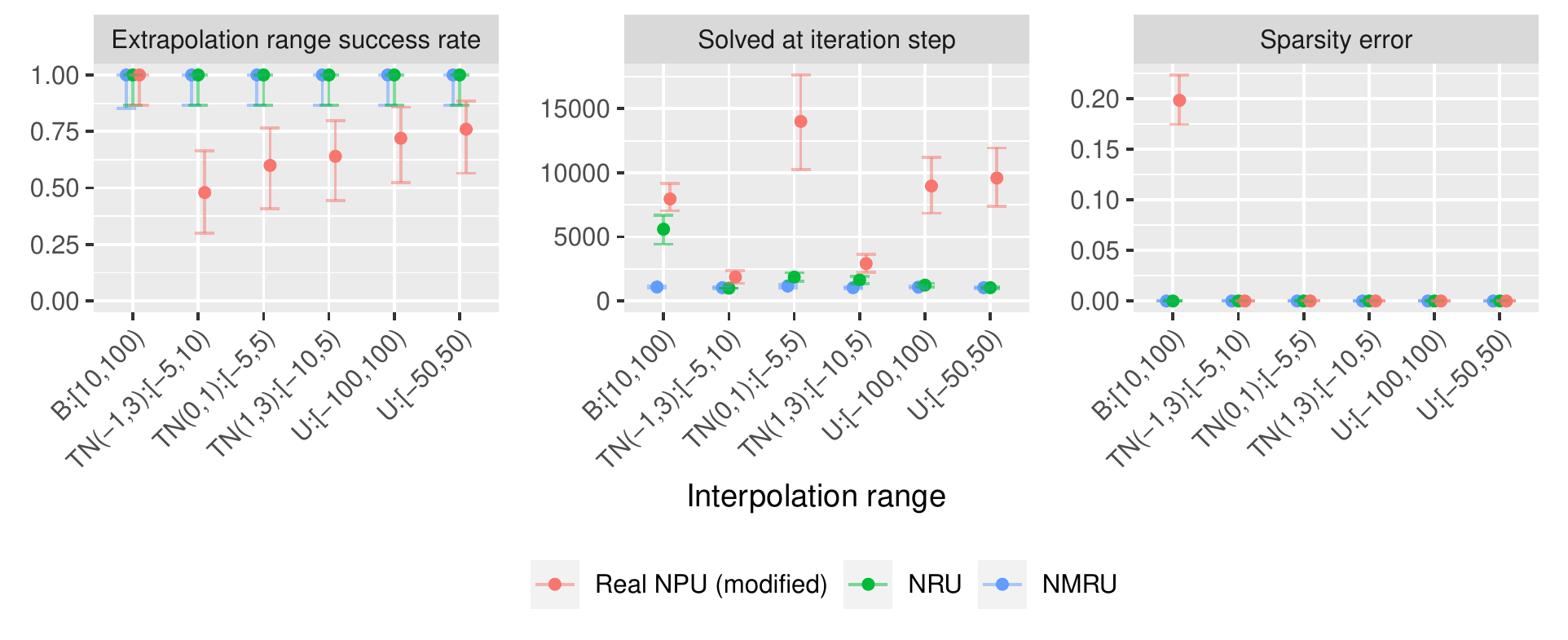}
  \caption{Division without redundancy (input size 2). Effect of the Benford, Truncated Normal and Uniform distribution.}
  \label{fig:rebuttal-in2}
\end{figure}

\begin{figure}[t]
  \centering
  \includegraphics[width=0.9\textwidth]{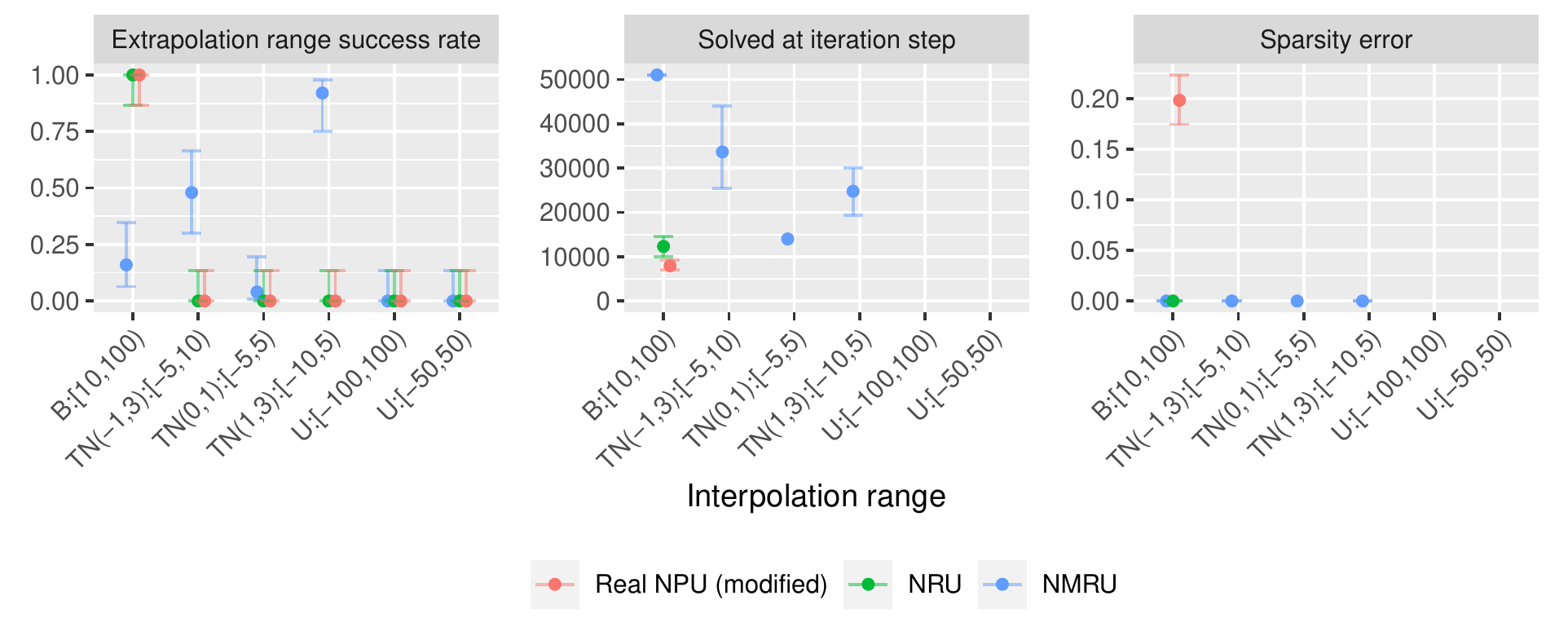}
  \caption{Division with redundancy (input size 10). Effect of the Benford, Truncated Normal and Uniform distribution.}
  \label{fig:rebuttal-in10}
\end{figure}

To discover more failure cases, we explore additional different distributions (i.e., Benford and Truncated Normal) and larger ranges around zero (Uniform distribution) defined in Table~\ref{tab:distribution-ranges}. 
Each distribution is tested on both the 2 input no redundancy (Figure~\ref{fig:rebuttal-in2}) and 10 input redundancy (Figure~\ref{fig:rebuttal-in10}) setting.

\begin{table}[t]
\centering
\caption{Interpolation (train/validation) and extrapolation (test) ranges used. Data (as floats) is drawn with the range values as the lower and upper bounds. TN = Truncated Normal distribution in the form TN(mean, sd)[lower bound, upper bound). B = Benford distribution. U = Uniform distribution.}
\vspace{1em}
\label{tab:distribution-ranges}
\begin{tabular}{lllll}
\toprule
\textbf{Interpolation} & {TN(-1, 3)[}-5, 10) & {TN(0,1)[}-5, 5) & {TN(1, 3)[}-10, 5) \\
\textbf{Extrapolation} & {TN(-10, 3)[}-15, -5) & {TN(10,1)[}5, 15) & {TN(10, 3)[}5, 15) \\\midrule
\textbf{Interpolation} & {B[}10, 100)   & {U[}-100, 100) & {U[}-50, 50) \\
\textbf{Extrapolation} & {B[}100, 1000) & {U[} -200, -100) $\cup$ [100, 200)] & {U[}[-100, -50) $\cup$ [50, 100)]   \\\bottomrule
\end{tabular}
\end{table}

\textbf{Uniform distributions:} All modules find the larger ranges to be challenging when redundant inputs exist. 
On the 2-input setup, both NRU and NMRU have full success, while the RealNPU (modified) has failure cases for both Uniform distributions (with success rates of 0.72 on \uniform[-100,100) and 0.76 on \uniform[-50,50)). 
On the 10-input size setup, all modules fail for all runs for both ranges. 

\textbf{Benford distribution:} This follows a more natural distribution compared to the Uniform distribution, known to underlie real world data such as accounting data \citep{hill1995statistical}. A large range is sampled, showing full success for the NRU and modified RealNPU on the 10 input setting. This implies the underlying issue from the failures of the Uniform distributions are attributed to the used of mixed signed inputs rather than the large ranges. However, the NMRU shows majority failures (failure rate 0.84) suggesting that large ranges are also an area of struggle for the module.  

\textbf{Truncated Normal distribution:} We discover further failure cases for the modules (especially the RealNPU (modified)). 
When trained using the 2-input setup, both NRU and SignNMRU have full success but the RealNPU (modified) has failure cases for all three distributions (with success rates 0.48, 0.6, 0.64 respectively).
When trained using the 10-input setup, both the NRU and RealNPU (modified) have no success in any range. 
The NMRU's success rate greatly varies depending on the range (being 0.48, 0.04 and 0.92 for TN(-1, 3)[-5, 10), TN(0,1)[-5, 5) and TN(1, 3)[-10, 5) respectively). This suggests that the NMRU works better when a majority of the inputs are likely to have the same sign and struggles with values around zero. 

\section{Discussion}\label{sec:discussion}
In this paper, we demonstrate the limitations of intepretable neural networks in learning to divide.
Using the no redundancy setting (size 2), we find that the Real NPU is challenged when training data consists of mixed-signed inputs even with our applied improvements. 
Increasing the difficulty to have an input redundancy (with 8 redundant and 2 relevant input values) magnifies this issue, but also introduces failure modes for the NRU and NMRU for negative ranges. 
The NRU is unable to handle any negative ranges, in which we conclude it is not wise to use with MSE. 
Alternate losses can improve certain failure cases though sometimes at the cost of performance on other ranges. For further details see Appendix~\ref{app:sltr-in10-losses} which displays results on a correlation and scale-invariant based loss. 
From training on different distributions, we find all modules struggle with the large Uniform distribution and Truncated Normal distributions a redundancy setting.

Our NMRU is the only module with reasonable success over all tested Uniform distributions in Section~\ref{sec:results}, but is challenged by extremely large ranges (see Figure~\ref{fig:rebuttal-in10}). The NMRU requires only $2I\times O$ learnable parameters, however this comes at the cost of the simplicity of the solution due to its exploitation of the identity rule; an issue the Real NPU does not have. 
\begin{figure}
\centering
\subfloat[Real NPU]{\includegraphics[width=4cm]{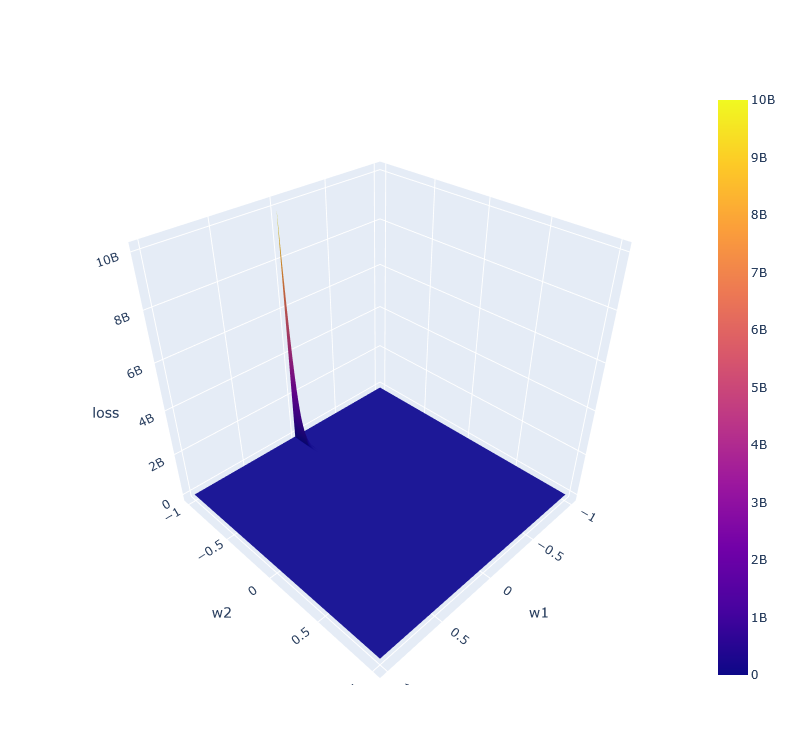}\label{subfig:loss-nau-realnpu}} 
\hspace{0.8cm}%
\subfloat[NRU]{\includegraphics[width=4cm]{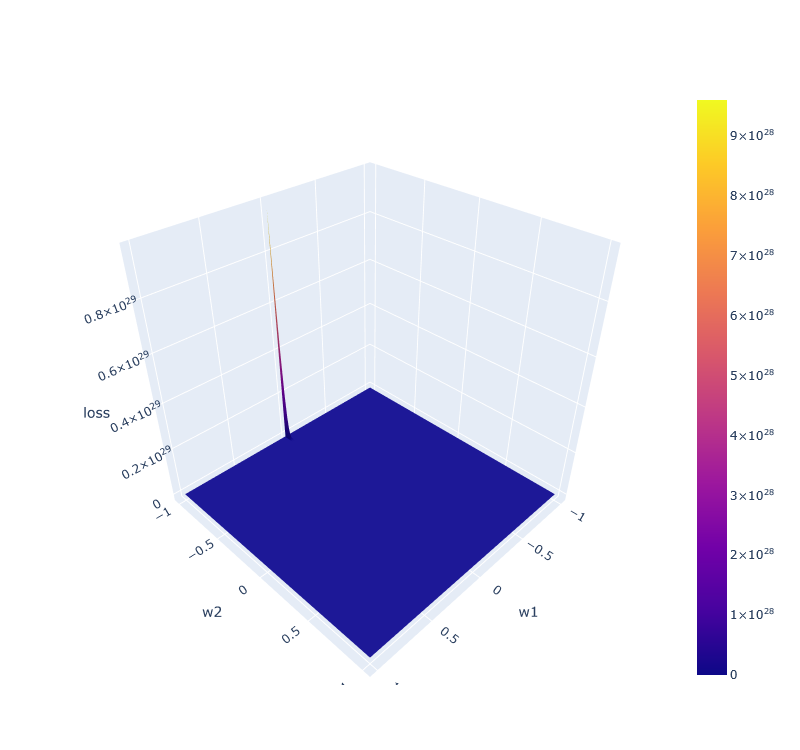}\label{subfig:loss-nau-nru}}   
\hspace{0.8cm}%
\subfloat[NMRU]{\includegraphics[width=4cm]{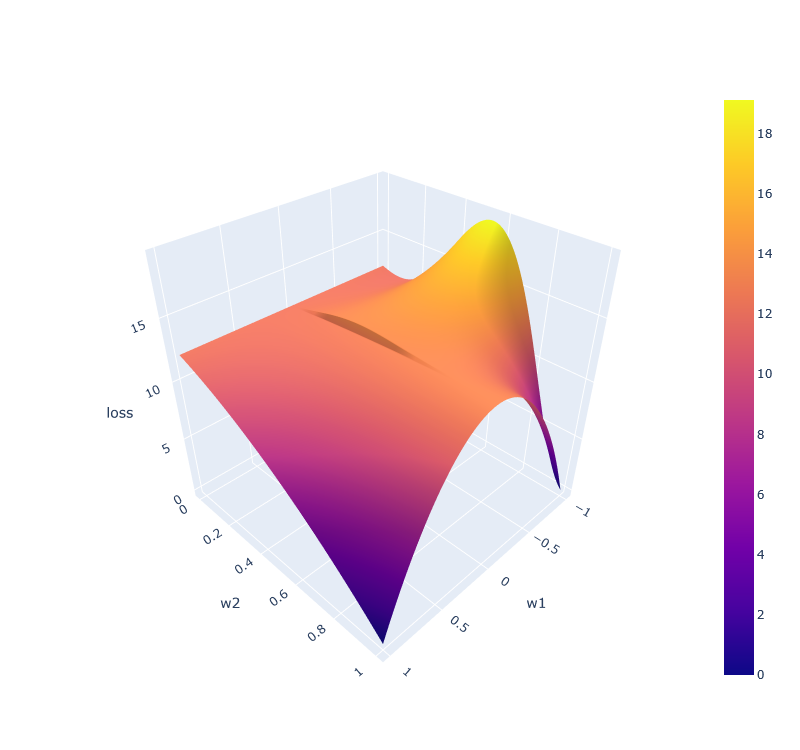}\label{subfig:loss-nau-nmru}}
\caption{Root Mean Squared loss curvature for the $\mathrm{NAU}$ stacked with either a $\mathrm{Real NPU}$, $\mathrm{NRU}$, or $\mathrm{NMRU}$. "The weight matrices are constrained to 
$\mathbf{W}_1 = \left[\protect\begin{smallmatrix}
w_1 & w_1 & 0 & 0 \\
w_1 & w_1 & w_1 & w_1
\protect\end{smallmatrix}\right]$, 
$\mathbf{W}_2 = \left[\protect\begin{smallmatrix}
w_2 & w_2
\protect\end{smallmatrix}\right]$. 
The problem is $(x_1 + x_2) \cdot (x_1 + x_2 + x_3 + x_4)$ for $x = \left(1, 1.2, 1.8, 2\right)$" 
\citep{madsen2020neural}. The ideal solution is $w_1 = w_2 = 1$, though other valid solutions do exist e.g., $w_1=-1, w_2=1$. (The NMRU's weight matrix would be 
$\mathbf{W}_2 = \left[\protect\begin{smallmatrix}
w_2 & w_2 & 0 & 0
\protect\end{smallmatrix}\right]$, 
and the Real NPU's 
$\mathbf{g} = \left[\protect\begin{smallmatrix}
1 & 1
\protect\end{smallmatrix}\right]$.
)
}
\label{fig:2L-losses}
\end{figure}
Once robust modules are attainable in a single layer setting, the next step would be to question performance when learning stacked modules, e.g. learning a stacked additive and multiplicative module. 
Previously, \citet[Figure~2]{madsen2020neural} illustrates the troubles for multiplicative models with the capacity for division. 
They show how a stacked summative-multiplicative module can lead to an exploding loss when the output of the summative module is close to 0 and the multiplicative model tries to divide.  
In Figure~\ref{fig:2L-losses}, we recreate their setup to produce the loss surfaces for the NAU-Real NPU\footnote{The NAU is a summative module \citep{madsen2020neural}.}, NAU-NRU and NAU-NMRU respectively.\footnote{Appendix~\ref{app:rmse-3d-plots} displays larger versions of these plots.} 
We find a similar issue with the Real-NPU and NRU, as both these units use a weight range of [-1,1]. 
In contrast, the NMRU, whose weight's range is limited to [0,1] does not have exploding losses.

In conclusion, division remains a challenge to learn using intepretable neural networks, even for the simplest tasks. 
Nevertheless, by identifying the specific areas causing difficulty (e.g., training ranges), and useful architecture properties (e.g., using a sign retrieval mechanism), we hope the community has better intuition for dealing with division and develop more robust modules to learn division. 

\bibliographystyle{plainnat}
\bibliography{references}
\appendix
\newpage
\section{Properties of a Division Module} \label{app:division-module-properties}
When building a division module, the following properties should be included:

\textbf{Ability to multiply:} Without multiplication the module is limited to expressing reciprocals. 

\textbf{Intepretable weights:} A good division module should produce generalisable solutions to out-of-bounds data. Using interpretable weights to represent exact operations is one way of doing so, e.g., -1 to divide, 1 to multiply, 0 to not select. For the scope of this paper we focus on discrete weights, however fractional weights can also be considered interpretable. For example, the Real NPU can express $\frac{1}{\sqrt{x_i}}$ using a weight value of -0.5.

\textbf{Calculating the output:} This can be decomposed into three tasks: magnitude calculation, sign calculation and input selection.

\textbf{Magnitude calculation:} Refers to calculating the output value for a calculation. This is achieved using discrete weight parameters. For example, the Real NPU and NRU use a weight value of  -1 for calculating reciprocals of selected input and 1 for multiplication, while the NMRU uses 1 for selecting an input element resulting in either a multiplication or reciprocal depending on the weight's position index. 

\textbf{Sign of the output:} Calculating the sign value (1/-1) of the output can occur at an element level in which the sign is calculated for each intermediary value as each input element is being processed, or at the higher input level in which the sign is calculated separately for the magnitude and then applied once the final output magnitude is calculated. The NRU uses the prior method while the Real NPU and NMRU use the latter method. 
If an input is 0 or considered irrelevant then the output sign will be 1. 
(Ablation studies on the NMRU, Figure~\ref{fig:nmru-ablation}, suggest the latter option which separately calculates the sign to be more beneficial). 

The Real NPU and NMRU use the cosine function to calculate the final sign of the module's output neuron. 
Below shows the state diagram of how the sign value (i.e. the state) of the output would change depending on the inputs and relevant parameters being processed. 
We only consider the discrete parameters for simplicity. 
Both the Real NPU and NMRU use the same state diagram but have different conditions for a state transition to occur. 
\begin{center}
\begin{tikzpicture}[scale=0.2]
\tikzstyle{every node}+=[inner sep=0pt]
\draw [black] (22.9,-36.9) circle (3);
\draw (22.9,-36.9) node {$s=-1$};
\draw [black] (39.2,-36.9) circle (3);
\draw (39.2,-36.9) node {$s=1$};
\draw [black] (20.22,-38.223) arc (-36:-324:2.25);
\draw (15.65,-36.9) node [left] {$b(s)$};
\fill [black] (20.22,-35.58) -- (19.87,-34.7) -- (19.28,-35.51);
\draw [black] (41.88,-35.577) arc (144:-144:2.25);
\draw (46.45,-36.9) node [right] {$b(s)$};
\fill [black] (41.88,-38.22) -- (42.23,-39.1) -- (42.82,-38.29);
\draw [black] (36.2,-36.9) -- (25.9,-36.9);
\fill [black] (25.9,-36.9) -- (26.7,-37.4) -- (26.7,-36.4);
\draw (31.05,-36.4) node [above] {$a(s)$};
\draw [black] (25.9,-36.9) -- (36.2,-36.9);
\fill [black] (36.2,-36.9) -- (35.4,-36.4) -- (35.4,-37.4);
\draw [black] (39.2,-31.3) -- (39.2,-33.9);
\fill [black] (39.2,-33.9) -- (39.7,-33.1) -- (38.7,-33.1);
\end{tikzpicture}
\end{center}

The conditions for the Real NPU transition functions $a(s)=-s$ and $b(s)=s$, where $s$ is the state value -1, or 1, are defined as follows: 
\begin{align*}
a(s):& x_i < 0 \wedge w_{i,o} \in \{-1,1\} \wedge g_i = 1\;,\\
b(s):& x_i \geq 0 \lor w_{i,o} = 0 \lor g_i = 0\;.
\end{align*}
Transitioning from one sign to another only occurs if the input element ($x_i$) is negative and is considered relevant i.e. the gate ($g_i$) and weight value ($w_{i,o}$) is non-0. 
In contrast, to remain at a state requires either the input element to be $\geq 0$ or not be considered relevant.

The conditions for the NMRU transition functions $a(s)=-s$ and $b(s)=s$, where $s$ is the state value -1, or 1, are defined as follows: 
\begin{align*}
a(s):& x_i < 0 \wedge w_{i,o} = 1\;,\\
b(s):& x_i \geq 0 \lor w_{i,o} = 0\;.
\end{align*}
Transitioning from one sign to another only occurs if the input element ($x_i$) is negative and is considered relevant i.e. the  weight value ($w_{i,o}$) is 1. 
To remain at a state requires either the input element to be $\geq 0$ or the weight value to not select the input.

\textbf{Selection:} Not all inputs are relevant for the output value. To process any irrelevant input elements can be interpreted as converting to the identity value of multiplication/division (=1).
The identity property means that any value multiplied/divided by the identity value remains at the original number.
Hence, irrelevant inputs are converted into 1 (rather than being masked out to 0). 
For the multiplication case, this stops the output becoming 0, and for division it avoids the divide by 0 case. 
For all the explored modules, a weight value of 0 will deal with the irrelevant input case. However, the Real NPU goes a step further by also having an additional gate vector with the purpose of learning to select relevant inputs. 
Such gating has been proven to be helpful for an NPU based module \citep{heim2020neural}, but may not be necessary when dealing with weights between [0,1] like in the NRMU (see Appendix~\ref{app:nmru-extra-exps}).

\newpage
\section{Neural Addition and Neural Multiplication Units' (NAU \& NMU)}\label{app:nau-nmu}
\citet{madsen2020neural} develop two modules: one for dealing with addition and subtraction (the NAU) and the other for multiplication (the NMU). 
NAU output element $a_o$ is defined as 
\begin{align}
\textrm{NAU}: a_o &= \sum_{i=1}^{I} \left(W_{i,o}\cdot \mathrm{x}_{i} \right) \label{eq:nau}
\end{align}
where $I$ is the number of inputs. 
The NMU output element $m_o$ is defined as  
\begin{align}
\textrm{NMU}: m_o &= \prod_{i=1}^{I} \left(W_{i,o}\cdot \mathrm{x}_{i} + 1 - W_{i,o} \right) . \label{eq:nmu}
\end{align}
Before passing a input through a module, the weight matrix is clamped to [-1,1] for the NAU or [0,1] for the NMU. 
Weights are ideally discrete values, where the NAU is 0, 1, or -1, representing no selection, addition and subtraction, and the NMU is 0 or 1, representing no selection and multiplication. 
To enforce discretisation of weights both units have a regularisation penalty for a given period of training. The penalty is
\begin{equation}
\lambda \cdot \frac{1}{I \cdot O} \sum_{o=1}^{O} \sum_{i=1}^{I} \min\left(|W_{i,o}|, 1 - |W_{i,o}|\right) ,
\end{equation}
where $O$ is the number of outputs and $\lambda$ is defined as
\begin{equation}
\lambda = \hat{\lambda} \cdot \max\left(\min\left(\frac{iteration_i - \lambda_{start}}{\lambda_{end} - \lambda_{start}}, 1\right), 0\right) .
\label{eq:regualizer-scaling}
\end{equation}
Regularisation strength is scaled by a predefined $\hat{\lambda}$. The regularisation will grow from 0 to $\hat{\lambda}$ between iterations $\lambda_{start}$ and $\lambda_{end}$, after which it plateaus and remains at $\hat{\lambda}$. 

\newpage
\section{Experiment Parameters}\label{app:parameters}
Tables~\ref{tab:exp-params} and \ref{tab:sltr-npu-params} for the breakdown of parameters used in the Single Module Tasks. 
Table~\ref{tab:mixed-sign-datasets} gives the interpolation and extrapolation ranges used in the mixed-sign datasets tasks. 

\begin{table}[h]
\caption{Parameters which are applied to all modules. Parameters have been split based on the experiment. $^*$Validation and test datasets generate one batch of samples at the start which gets used for evaluation for all iterations. $^{\dagger}$ the Real NPU modules use a value of 1.}
\label{tab:exp-params}
\vskip 0.1in
\begin{center}
    \begin{tabular}{p{3cm}p{4.5cm}p{4.5cm}}
    \toprule
    \textbf{Parameter}          & \textbf{Without redundancy} & \textbf{With redundancy}     \\ \midrule
    \textbf{Layers}             & 1                                & 1                              \\ 
    \textbf{Input size}         & 2                              & 10                              \\ 
    \textbf{Total iterations}   & 50,000                        & 100,000                          \\ 
    \textbf{Train samples}      & 128 per batch                    & 128 per batch                  \\ 
    \textbf{Validation samples$^*$} & 10000                            & 10000                          \\ 
    \textbf{Test samples$^*$}       & 10000                            & 10000                          \\ 
    \textbf{Seeds}              & 25                               & 25                             \\ 
    \textbf{Optimiser}          & Adam (with default parameters)   & Adam (with default parameters) \\ 
    \textbf{$\hat{\lambda}^{\dagger}$}     & 10                            & 10     \\ \bottomrule
    \end{tabular}
\end{center}
\vskip -0.1in
\end{table}

\begin{table}[h]
\centering
\caption{Parameters specific to the Real NPU modules for the Single Module Tasks.}
\label{tab:sltr-npu-params}
\vskip 0.1in
\begin{tabular}{ll}
\toprule
\textbf{Parameter}                  & \textbf{Value}    \\ \midrule
  ($\beta_{start}$,$\beta_{end}$)   & (1e-9,1e-7)    \\ 
  $\beta_{growth}$                  & 10             \\  
  $\beta_{step}$                    & 10000          \\ 
  $\hat{\lambda}$                   & 1              \\ \bottomrule
\end{tabular}
\vskip -0.1in
\end{table}

\begin{table}[h]
\caption{Mixed-Sign Datasets: The interpolation and extrapolation ranges to sample the two input elements for a single data sample. The target expression to learn is: input 1 $\div$ input 2.}
\label{tab:mixed-sign-datasets}
\centering
\vskip 0.1in
\begin{small}
\begin{sc}
    \begin{tabular}{lllll}
    \toprule
    & \multicolumn{2}{c}{Interpolation} & \multicolumn{2}{c}{Extrapolation} \\\cmidrule(lr){2-3}\cmidrule(lr){4-5}
    \textbf{Dataset} & \textbf{Input 1}   & \textbf{Input 2} & \textbf{Input 1}   & \textbf{Input 2} \\ \midrule
    1 & U{[}-2, -0.1)   & U{[}0.1, 2)   & U{[}-6, -2)   & U{[}2, 6) 
    \\ 
    2 & U{[}-2, -1)     & U{[}1, 2)     & U{[}-6, -2)   & U{[}2, 6)       
    \\ 
    3 & U{[}-2, 2)      & U{[}-2, 2)    & U{[}-6, -2)   & U{[}2, 6)        
    \\ 
    4 & U{[}0.1, 2)     & U{[}-2, -0.1) & U{[}2, 6)   & U{[}-6, -2)    
    \\ 
    5 & U{[}1, -2)      & U{[}-2, -1)  & U{[}2, 6)   & U{[}-6, -2)      
    \\ \bottomrule
    \end{tabular}
\end{sc}
\end{small}
\end{table}

\subsection{Parameter Initialisation}
We give the initialisations used on the different module parameters:

\textbf{Real NPU}: The real weight matrix uses the Pytorch's Xavier Uniform initialisation. The gate vector initialises all values to 0.5. (This is the same initialisation used in \citet{heim2020neural}.) 

\textbf{NPU}: The imaginary weight matrix is initialised to 0. The rest of the parameters are initialised same as the Real NPU. (This is the same initialisation used in \citet{heim2020neural}.)

\textbf{NRU}: The weight matrix uses a Xavier Uniform initialisation which can have a maximum range between -0.5 to 0.5 (depending on the network sizes). (This is the same initialisation the Neural Addition Unit uses \citep{madsen2020neural}.)

\textbf{NMRU}: The weight matrix uses a Uniform initialisation which can have a maximum range between 0.25 to 0.75 (depending on the network sizes). (This is the same initialisation the Neural Multiplication unit uses \citep{madsen2020neural}.)

\newpage
\section{Hardware and Time to Run Experiments}\label{app:hardware-and-timings}
All experiments were trained on the CPU, as training on GPUs takes considerably longer. 
All Real NPU experiments were run on \iridis (the University of \soton's supercomputer), where a compute node has 40 CPUs with 192 GB of DDR4 memory which uses dual 2.0 GHz Intel Skylake processors.
All NRU and NMRU experiments were run on a 16 core CPU server with 125 GB memory 1.2 GHz processors.

Table~\ref{tab:timings} displays time taken for each experiment to run a single seed for a single range. 
Timings are based on a single run rather than the runtime of a script execution because the queuing time from jobs when executing scripts is not relevant to the experiment timings. 
For a single model, a single experiment would have 225 runs (for 9 training ranges and 25 seeds).

\begin{table}[h]
\centering
\caption{Timings of experiments.}
\label{tab:timings}
\vskip 0.1in
\begin{tabular}{lll}
\toprule
\textbf{Experiment}     & \textbf{Model}    & \textbf{Approximate time for completing 1 seed (mm:ss)} \\ \midrule
\multirow{3}{*}{No redundancy (size 2)}    & Real NPU & 03:20    \\  
                                           & NRU      & 02:00    \\ 
                                           & NMRU     & 03:00    \\ \midrule
\multirow{3}{*}{With redundancy (size 10)} & Real NPU & 05:30    \\ 
                                           & NRU      & 05:00    \\
                                           & NMRU     & 05:15    \\ \bottomrule
\end{tabular}
\end{table}

\newpage
\section{NRU on the Single Module Task (no redundancy): Effect of Learning Rate}\label{app:nru-in2-lr}
\begin{figure}[h]
  \centering
  \includegraphics[width=0.9\textwidth]{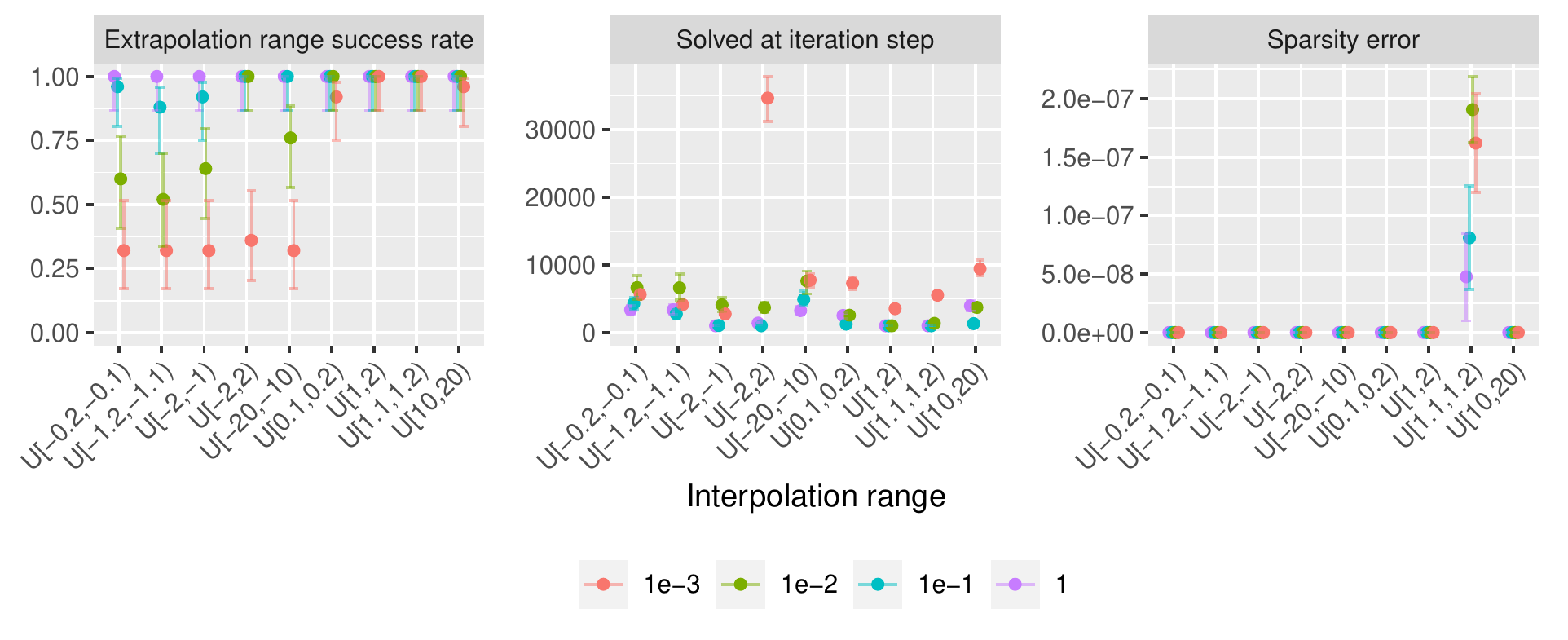}
  \caption{Different learning rates on the NRU for the Single Module Task (no redundancy)}
  \label{fig:nru-in2-lr}
\end{figure}

Figure~\ref{fig:nru-in2-lr} displays the effect of different learning rates for the NRU. An learning rate of 1 gets full success on all ranges with performance deteriorating as the learning rate reduces. 

\newpage
\section{Division by small values: Experimental Results} \label{app:divBy0}
This section shows the results on trying to learn the reciprocal/division of values close to zero using the Real NPU, NRU and NMRU.
We train and test on the ranges where the lowest bound is 0 and the upper bounds are: 1e-4, 1e-3, 1e-2, 1e-1 and 1. 
Unless stated otherwise, the hyperparameters of a model are set to what is used for the Single Layer Task without redundancy. 
The first task runs for 5,000 iterations with no regularisation for any module.  
The second and third tasks both run for 50,000 iterations.

Due to precision errors, a solution with the ideal parameters will not evaluate to a MSE of 0. 
Therefore, we calculate thresholds which the test MSE should be within. 
A threshold value for a task is calculated from evaluating the MSE of each range's test dataset for each module, using the `golden' weight values and adding an epsilon term\footnote{The term is the pytorch default eps value, torch.finfo().eps} to the resulting error which takes into account precision errors. 
All experiments are run using 32-bit precision. 

\begin{figure}[h]
  \centering
  \includegraphics[width=0.9\textwidth]{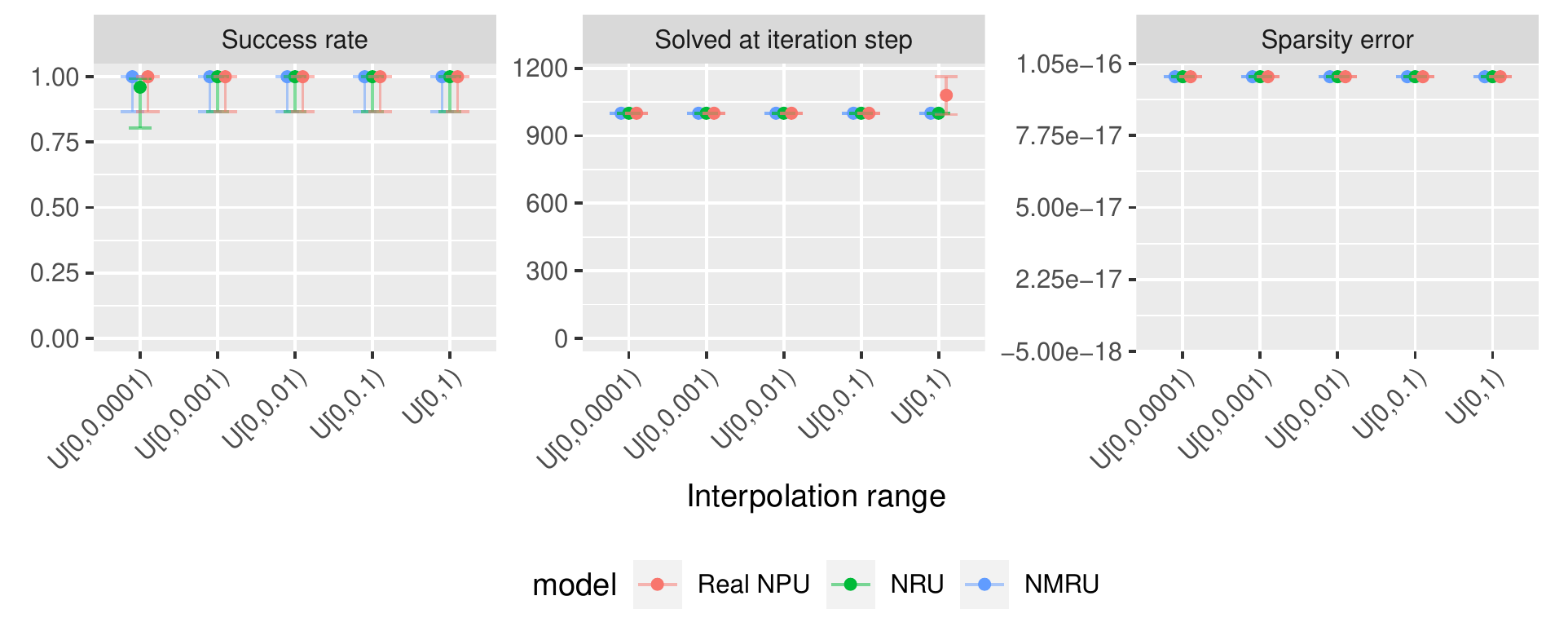}
  \caption{Input: [a], output $\frac{1}{a}$. Learns reciprocal when there is no input redundancy.}
  \label{fig:divBy0-easy}
\end{figure}

In general, successful runs take longer to solve as the input ranges become smaller. 
The simplest task, of taking the reciprocal when the input size is 1 (Figure~\ref{fig:divBy0-easy}) is achieved with ease for all modules, though for \uniform[0,1e-4), we find the NRU begins to start struggling.

\begin{figure}[h]
  \centering
  \includegraphics[width=0.9\textwidth]{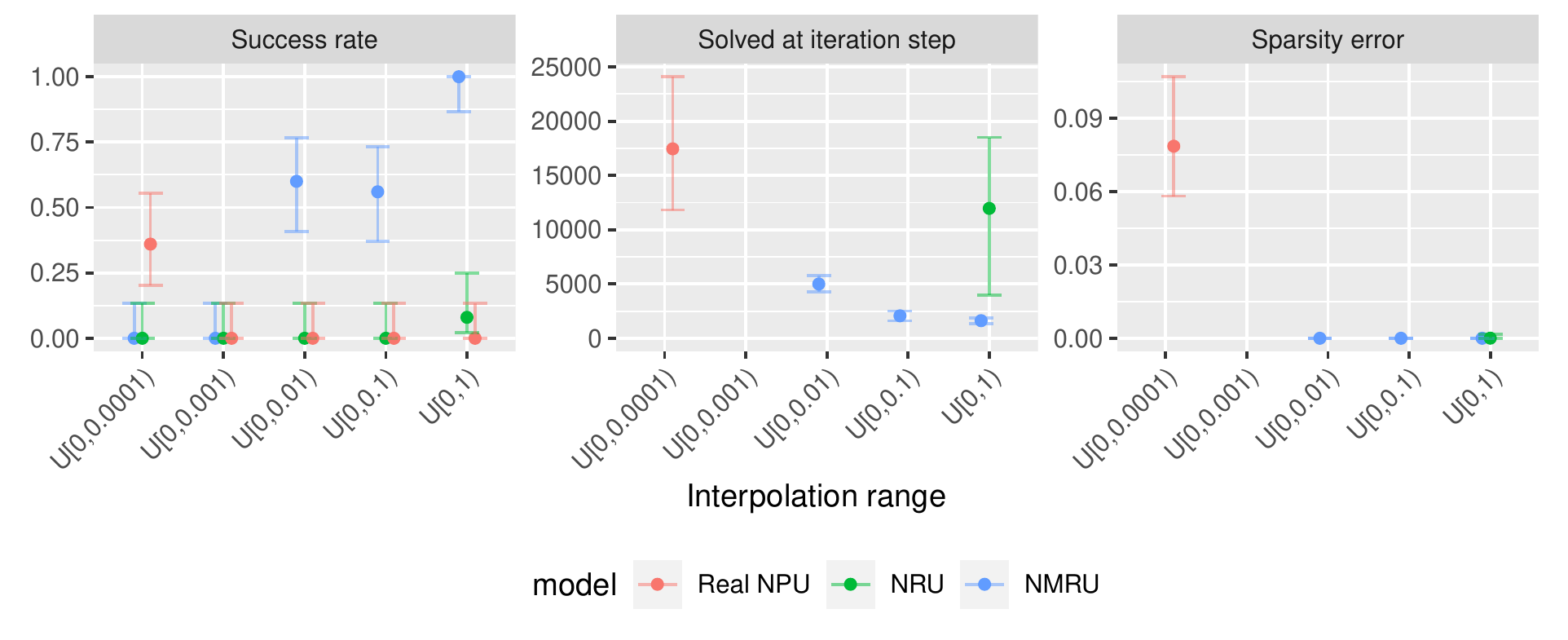}
  \caption{Input: [a,b], output $\frac{1}{a}$. Learns reciprocal of the first input when there is redundancy.}
  \label{fig:divBy0-medium}
\end{figure}

Introducing a redundant input (Figure~\ref{fig:divBy0-medium}) greatly impacts performance with only the NMRU able to achieve reasonable success for the larger ranges. The successes shown for the Real NPU at range \uniform[0, 1e-4) are false positives caused by the $\epsilon$ in the architecture used for stability. 
Test false positives can also be indicated by the high sparsity error of the weights. 

\begin{figure}[h]
  \centering
  \includegraphics[width=0.9\textwidth]{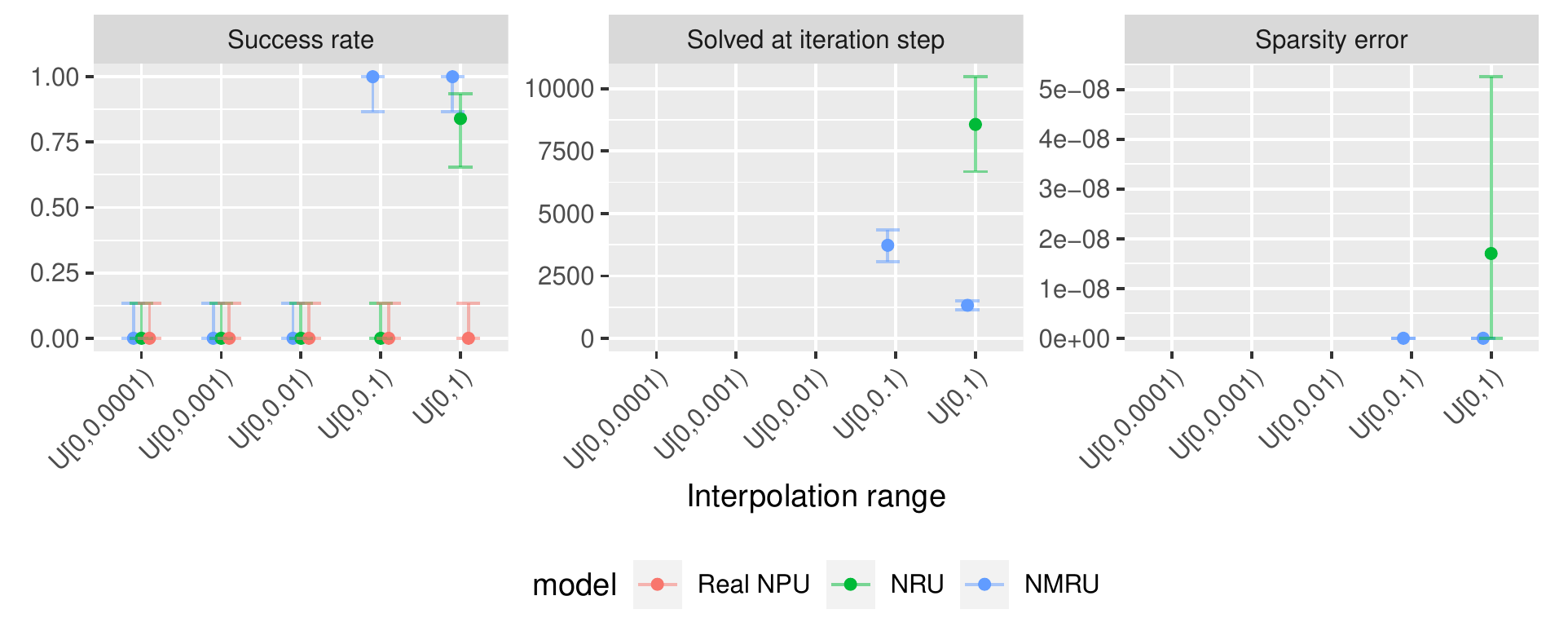}
  \caption{Input: [a,b], output $\frac{a}{b}$. Learns division of the first and second value when there is no redundancy.}
  \label{fig:divBy0-hard}
\end{figure}

Modifying the task to division (Figure~\ref{fig:divBy0-hard}), meaning the redundant input is now relevant, shows improvement for the NMRU and NRU for the larger ranges. 

\newpage
\section{Real NPU; Single Module Task (without Redundancy): Additional Experiments}\label{app:realnpu-in2-extra}
\begin{figure}[h]
  \centering
  \includegraphics[width=0.4\textwidth]{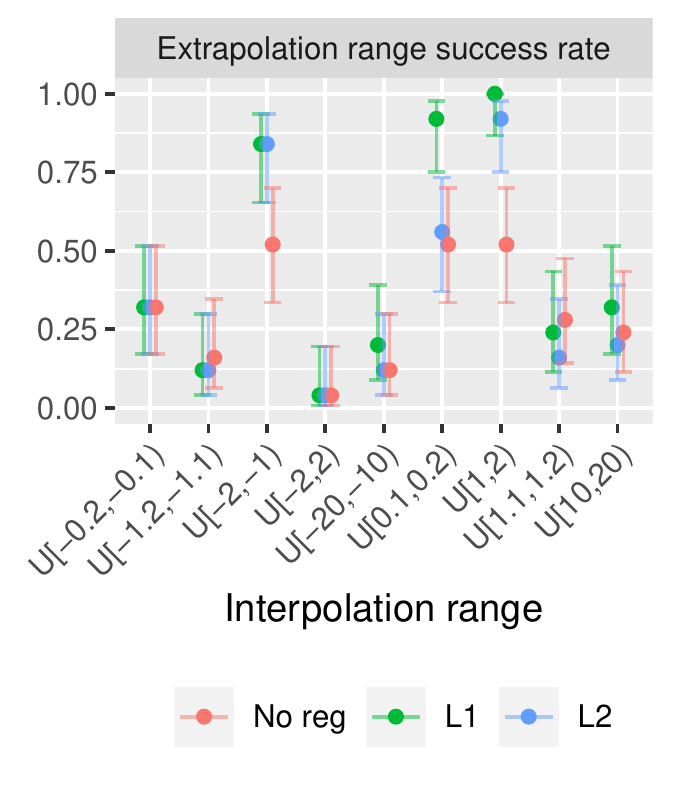}
  \caption{Applying no regularisation, L1 regularisation and L2 regularisation to enforce sparsity in weights.}
  \label{fig:sltr-in2-npu-reg}
\end{figure}

Figure~\ref{fig:sltr-in2-npu-reg} shows results of using the NPU for the 2-input task. 
Of the 9 tested ranges, L2 has a lower success rate than L1 for 5 ranges and has the same success rate for the remaining 4 ranges. If L2 regularisation is used instead of no regularisation, it performs worse in 3 (of the 9) ranges, better on 3 ranges and the same on the remaining 3 ranges.

\newpage
\section{Real NPU; Single Module Task (with Redundancy): Additional Experiments}\label{app:realnpu-in10-extra}
\begin{figure}[h]
  \centering
  \includegraphics[width=0.9\textwidth]{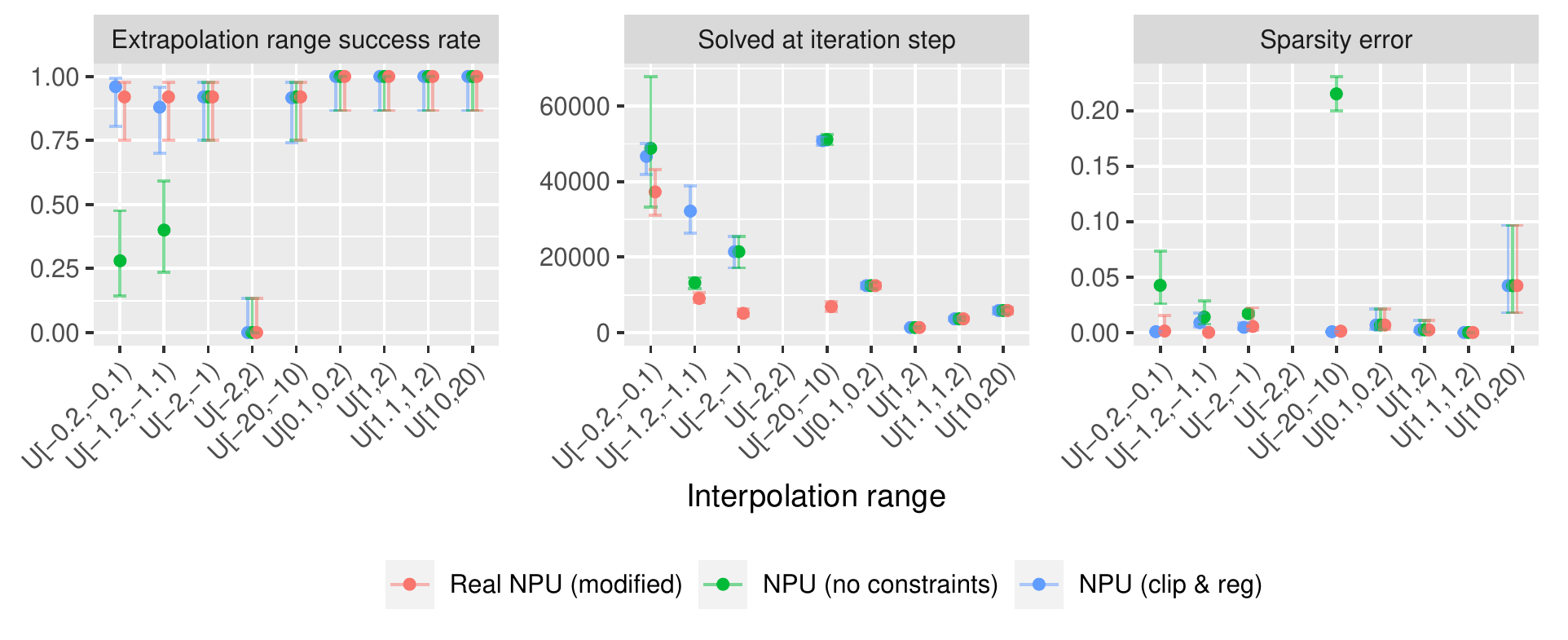}
  \caption{Adapting the Real NPU to use complex weights (NPU) on the Single Module Task with redundancy. Compares the NPU architecture with the Real NPU modifications (i.e. NPU (no constraints)) and the same model but with the imaginary weights clipped to [-1,1] and L1 sparsity regularisation on the complex weights (i.e. NPU (clip \& reg)).}
  \label{fig:sltr-in10-npu}
\end{figure}

Figure~\ref{fig:sltr-in10-npu} shows results of using the NPU for the task with redundancy.

\begin{figure}[h]
  \centering
  \includegraphics[width=0.9\textwidth]{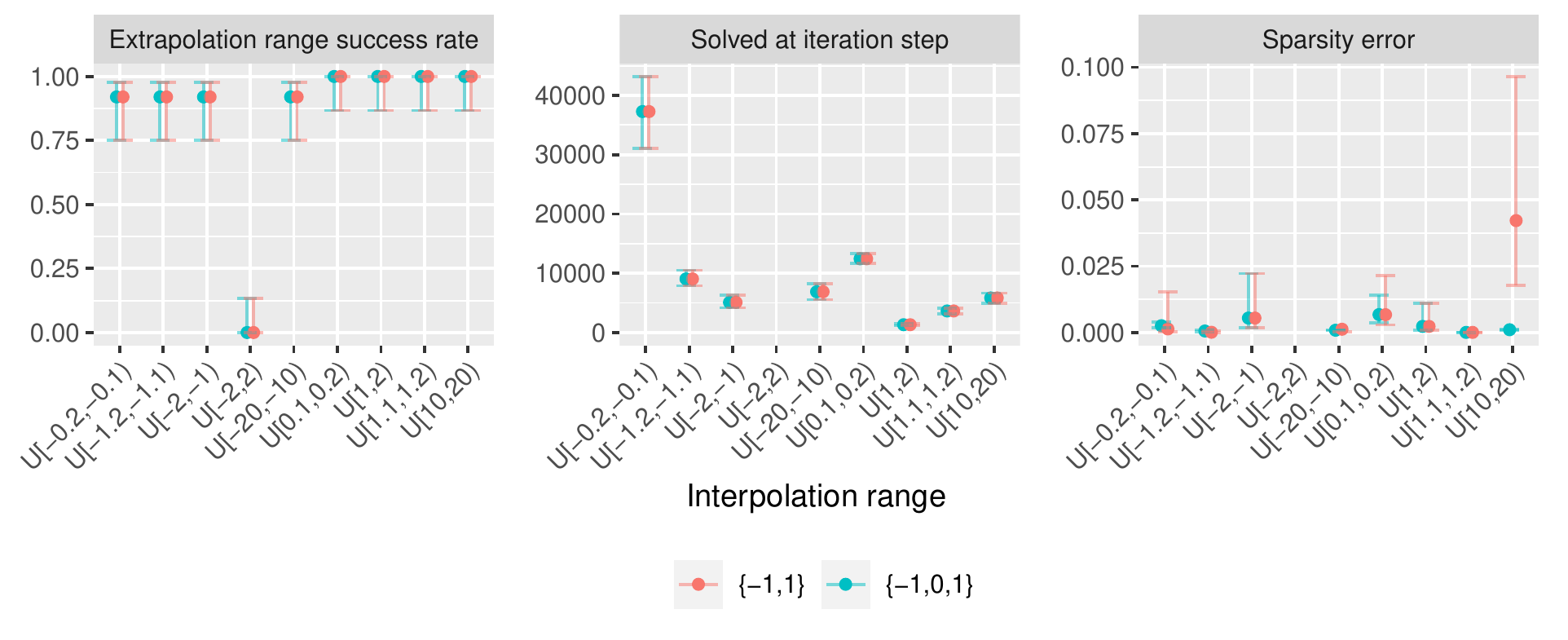}
  \caption{Comparing weight discretisation on the NPU weights which penalises not having weight of $\{-1,1\}$ vs $\{-1,0,1\}$.}
  \label{fig:realnpu-W-reg}
\end{figure}

Figure~\ref{fig:realnpu-W-reg} shows how modifying the weight discretisation to not penalise weights at 0 does not effect success.

\newpage
\section{NMRU; Single Module Task with Redundancy (Additional Experiments)}\label{app:nmru-extra-exps}
This section further explores the NMRU architecture. 

\begin{figure}[h]
  \centering
  \includegraphics[width=0.9\textwidth]{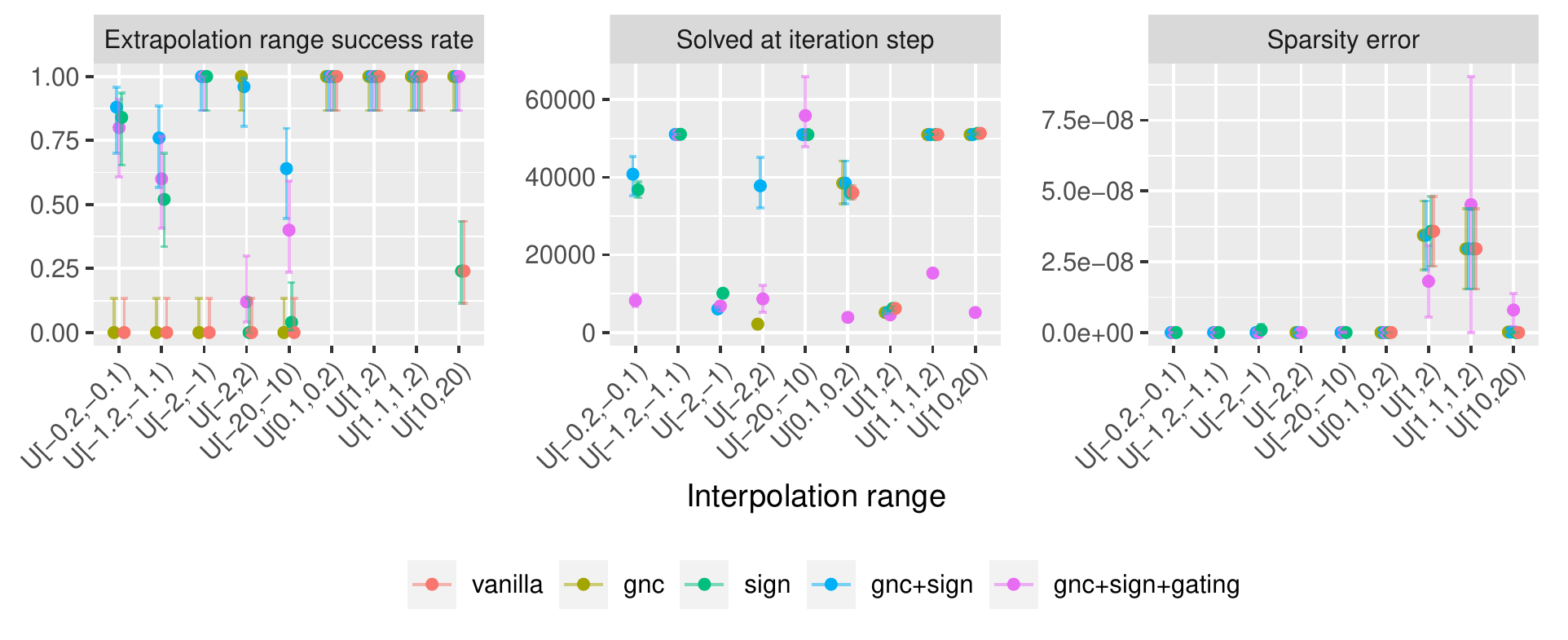}
  \caption{Ablation study for the NMRU.}
  \label{fig:nmru-ablation}
\end{figure}

Figure~\ref{fig:nmru-ablation} shows an ablation study on different components of the NMRU architecture. 
Removing both the sign retrieval and grad norm clipping performs poorly over a majority of ranges (including positive ranges). 
Gradient norm clipping alone is unable to solve the issue in learning negative ranges, however fully succeeds on the \uniform[-2,2) range. 
Using the sign retrieval without the gradient clipping gains successes for the negative ranges, though performance on \uniform[2,-2) is effected. 
However, including both gradient clipping and sign retrieval results in separating the calculation of the magnitude of the output and its sign while having reasonable gradients, gaining the most improvement over the vanilla NMRU. 
Further including a learnable gate vector (like the Real NPU), which is applied to the input vector, hinders performance. 
The largest solved at iteration step seems to be bounded at approximately 50,000 iterations which correlates to the point at which the sparsity regularisation begins, which highlights the importance of discritisation. 
Even with the different ablations, the sparsity errors of the successful seeds remain extremely low (which is not always the case for the Real NPU (see Figure~\ref{fig:sltr-in10})).

\begin{figure}[h]
  \centering
  \includegraphics[width=0.9\textwidth]{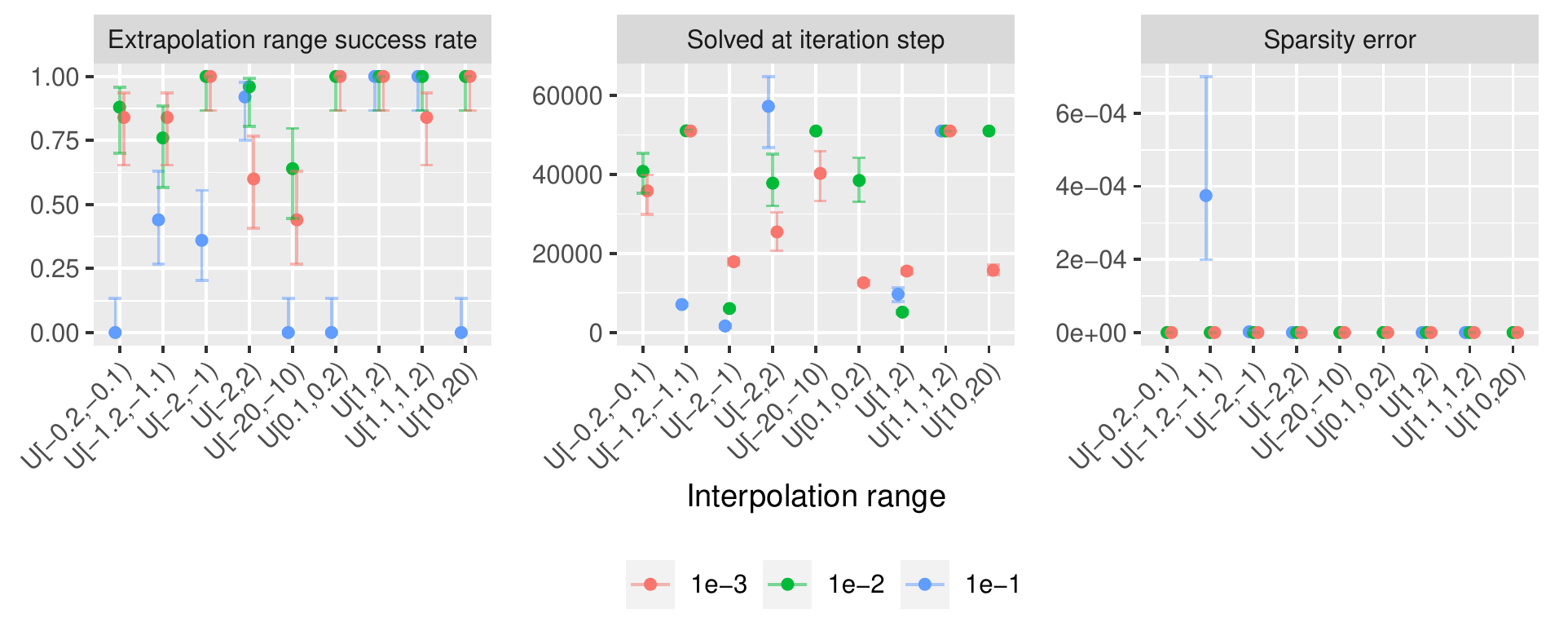}
  \caption{Effect of different learning rates on the NMRU}
  \label{fig:nmru-lr}
\end{figure}

Figure~\ref{fig:nmru-lr} shows the effect of using different learning rates on the NMRU (with grad norm clipping and sign retrieval) using an Adam optimiser. Too low a learning rate struggles on the mixed-sign range \uniform[-2,2). Too high a learning leads to no success on multiple ranges.  

\begin{figure}[h]
  \centering
  \includegraphics[width=0.9\textwidth]{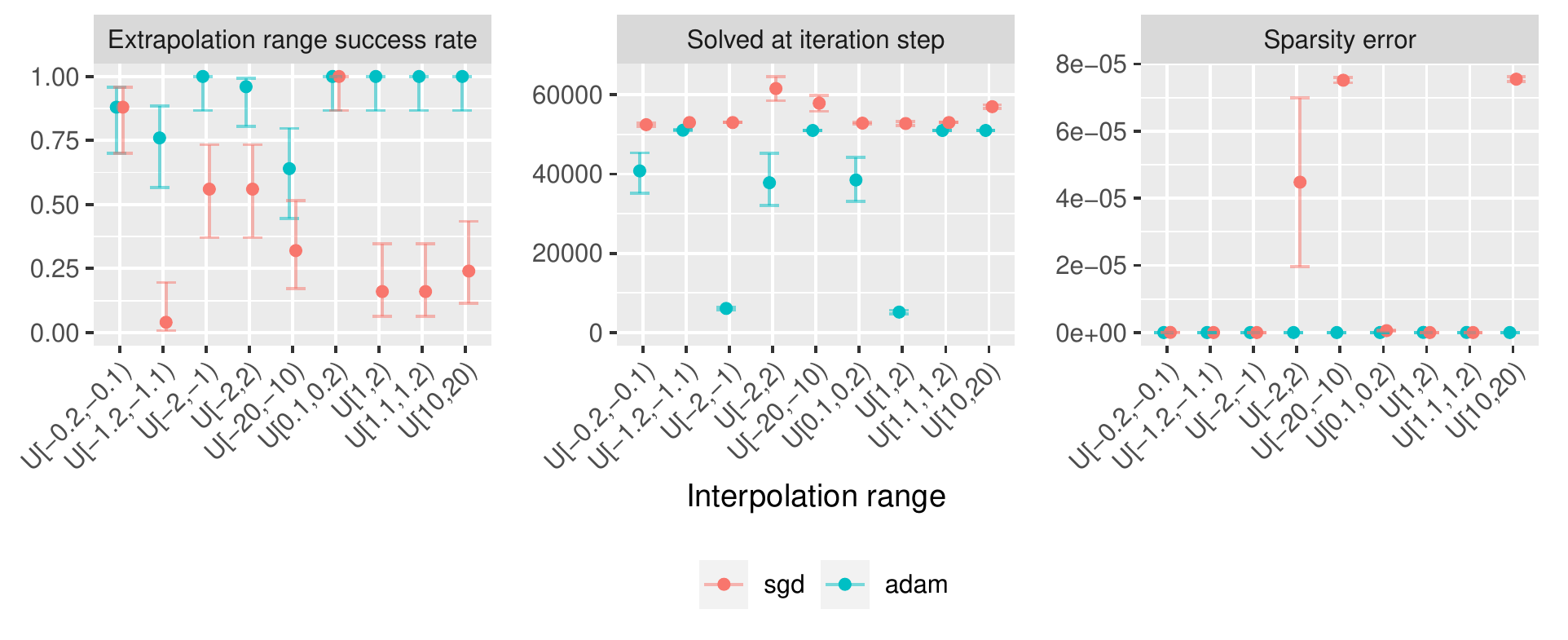}
  \caption{Effect of optimiser on the NMRU. SGD = Stochastic Gradient Descent.}
  \label{fig:nmru-optimiser}
\end{figure}

Figure~\ref{fig:nmru-optimiser} compares training the NMRU with either an Adam and SGD optimiser. 
As expected, Adam outperforms SGD in all ranges (except two, where both perform equally). 
This difference in performance can be accounted for by Adam's ability to scale the step size of each weight, which can compliment the clipped gradient norm of the NMRU, in contrast to the SGD's global step size.

\newpage
\section{NRU; Single Module Task (with Redundancy): Calculating the Sign Separately}\label{app:nru-sepSigns}
\begin{figure}[h]
  \centering
  \includegraphics[width=0.9\textwidth]{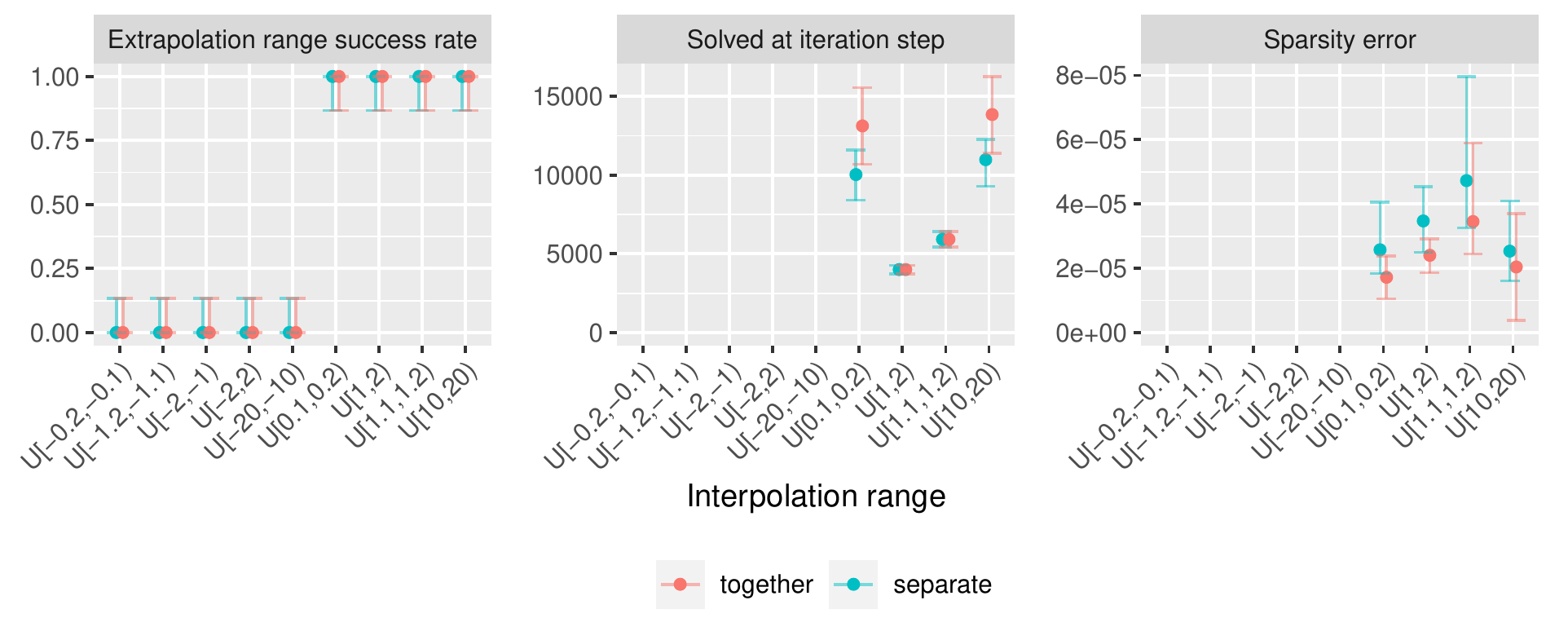}
  \caption{NRU on the redundancy experiment comparing a module which calculates the magnitude and sign together vs calculating the magnitude and sign separately and then combining them.}
  \label{fig:nru-in10-sepSigns}
\end{figure}

The `separate NRU' module calculates the magnitude and sign separately and then combines them using multiplication together once all input elements are accounted for. The following definition is used to calculate a NRU with separate magnitude and sign calculation, 
\begin{align}
z_o &= \prod_{i=1}^{I} \left(|\mathrm{x}_{i}|^{W_{i,o}}\cdot|W_{i,o}| + 1 - |W_{i,o}|\right) \cdot \prod_{i=1}^{I} \sign(\mathrm{x}_{i})^{\textrm{round}(W_{i,o})} \;.\label{eq:nru-sepSign}
\end{align}
Figure~\ref{fig:nru-in10-sepSigns} shows results, where the separate sign method shows no difference in success to the original NRU architecture.

\newpage
\section{Effect of Different Losses on the Single Module Task (with Redundancy)} \label{app:sltr-in10-losses}
\begin{table}[h]
\centering
\caption{The properties of different loss functions.}
\label{tab:loss-properties}
\vskip 0.1in
\begin{tabular}{llll}
\toprule
                         & MSE & PCC  & MAPE \\ \midrule
Batch mean      & \cmark       & \cmark        & \cmark        \\ 
Standardisation &              & \cmark        & \cmark        \\ 
Difference of prediction from target    & \cmark       &               & \cmark  \\ 
Projection      &              & \cmark        &               \\ 
Mean centering  &              & \cmark        &               \\ \bottomrule
\end{tabular}
\end{table}
Different losses induce different loss landscapes impacting the areas of success for a module. We explore the effects of three different losses including the MSE, Pearson's Correlation Coefficient (Equation~\ref{eq:pcc-loss}), and the Mean Absolute Precision Error (Equation~\ref{eq:mape-loss}). We use the division task with 10 inputs. 
The properties of each loss is summarised in Table~\ref{tab:loss-properties}.
All experiment parameters match the original MSE runs in the main experiments. The only difference is the loss used. 

\begin{equation}
\begin{aligned}
v_{x,i} &= (\hat{y}_i-\bar{\hat{y}}) ,\quad 
s_x = \sqrt{\textrm{clamp}(\frac{1}{N}\sum_i^N v_{x,i}^2 \textrm{, } \epsilon)} \\
v_{y,i} &= (y_i-\bar{y}) ,\quad 
s_y = \sqrt{\textrm{clamp}(\frac{1}{N}\sum_i^N v_{y,i}^2 \textrm{, } \epsilon)} \\
r &= \frac{1}{N}
    \sum_i^N(\frac{v_{x,i}}{s_x + \epsilon} \cdot           \frac{v_{y,i}}{s_y + \epsilon})
\end{aligned}
\end{equation}
\begin{equation}
\begin{aligned}
\pcc :=& 1 - r \label{eq:pcc-loss}
\end{aligned}
\end{equation}
where N is the batch size, and the means ($\bar{\hat{y}}$ and $\bar{y}$) are taken over the batch. $\epsilon$ is used to provide better numerical stability. 
\begin{equation}
\begin{aligned}
\mape :=& \frac{1}{N}\sum_i^N(\frac{|y_i - \hat{y}_i|}{y_i}) \label{eq:mape-loss}
\end{aligned}
\end{equation}

\begin{figure}[h]
  \centering
  \includegraphics[width=0.9\textwidth]{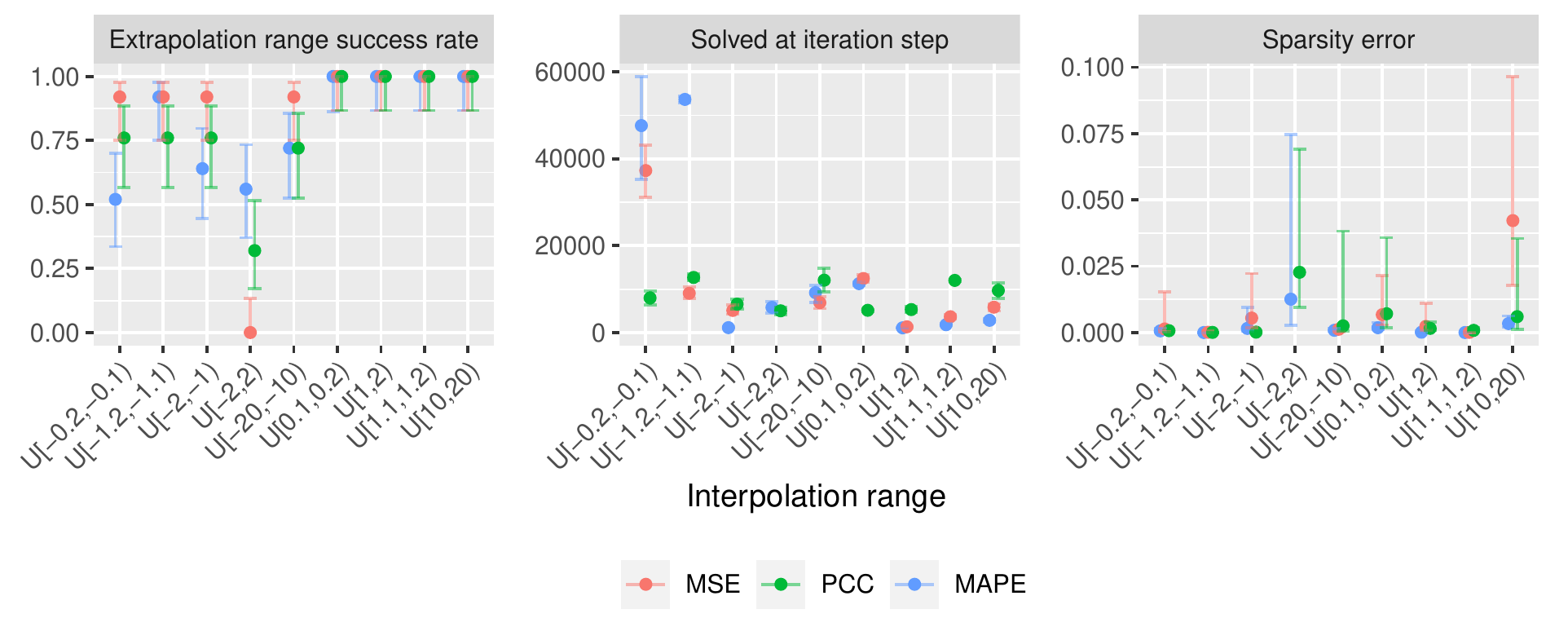}
  \caption{Single Module Task with redundancy on the Real NPU, comparing different loss functions.}
  \label{fig:losses-realnpu}
\end{figure}

\paragraph{Real NPU (Figure \ref{fig:losses-realnpu})} Both the Real NPU and MAPE are able to get success on the \uniform[-2,2) range, which the MSE completely fails on, implying that having a loss with standardisation is useful. 
However, in order to gain successes in the mixed-sign range, the other negative ranges have reduced in success for both PCC and MAPE. 
Both speed and sparsity retain similar performance to MSE in a majority of cases, with PCC solving especially fast for all tested ranges. 

\begin{figure}[h]
  \centering
  \includegraphics[width=0.9\textwidth]{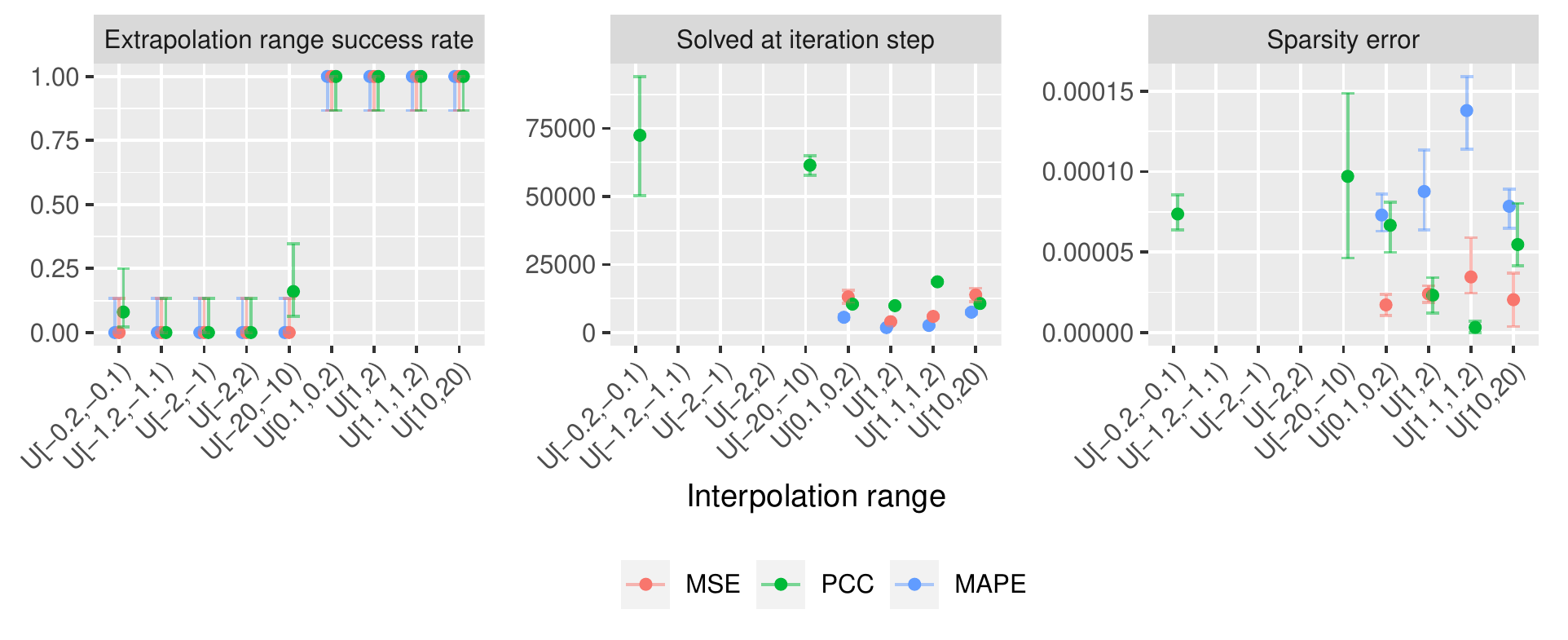}
  \caption{Single Module Task with redundancy on the NRU, comparing different loss functions.}
  \label{fig:losses-nru}
\end{figure}

\paragraph{NRU (Figure \ref{fig:losses-nru})}
Different losses have little effect on the NRU.
All three losses perform well on the positive ranges. 
Compared to the Real NPU, the PCC loss on the NRU takes longer to converge to a success for negative ranges.

\begin{figure}[h]
  \centering
  \includegraphics[width=0.9\textwidth]{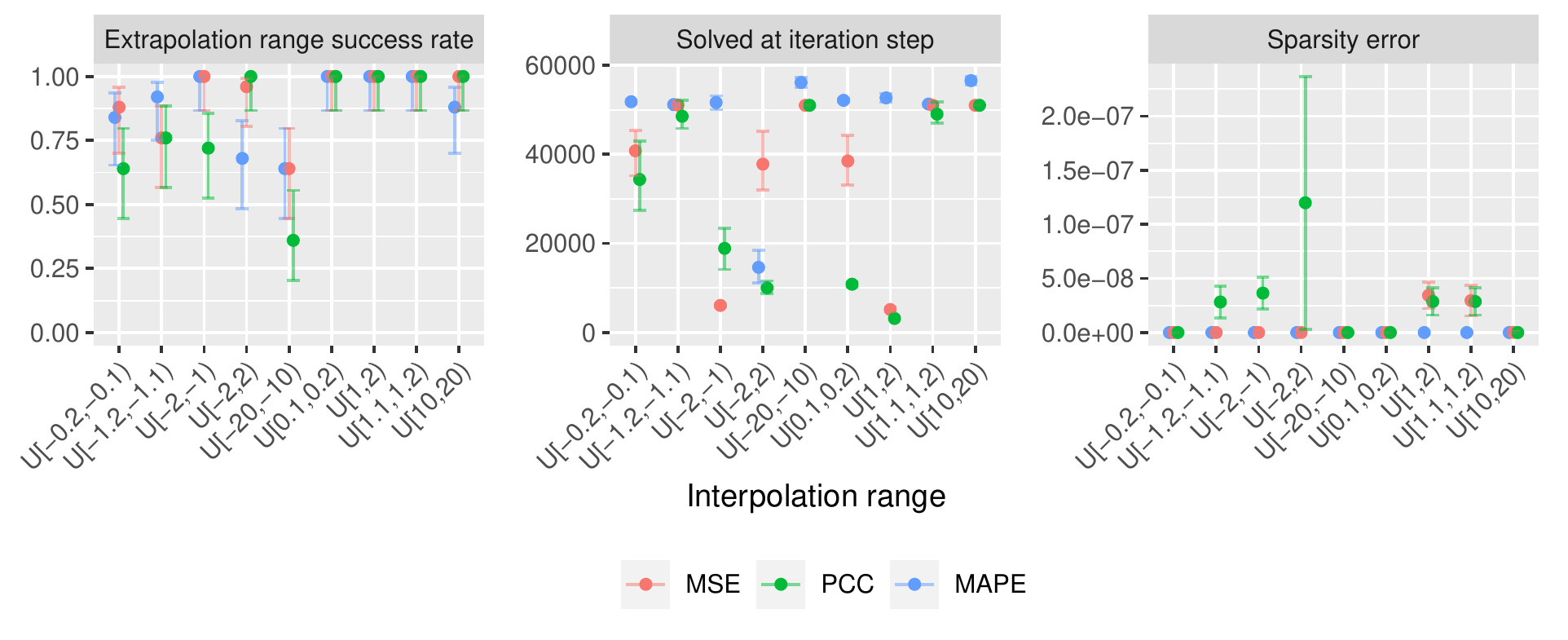}
  \caption{Single Module Task with redundancy on the NMRU, comparing different loss functions.}
  \label{fig:losses-nmru}
\end{figure}

\paragraph{NMRU (Figure \ref{fig:losses-nmru})}
All three loses perform reasonably well, with the PCC struggling the most. Unlike the other units, \uniform[-20,-10) causes the most trouble, whereas \uniform[-2,2) gains near to full success on two of the three losses.

\newpage
\section{RMSE Loss Landscapes} \label{app:rmse-3d-plots}
For clarity, we show bigger versions of each subplot from Figure~\ref{fig:2L-losses}. 
\begin{figure}[]
    \centering
    \begin{subfigure}{0.8\textwidth}
        \includegraphics[width=\linewidth]{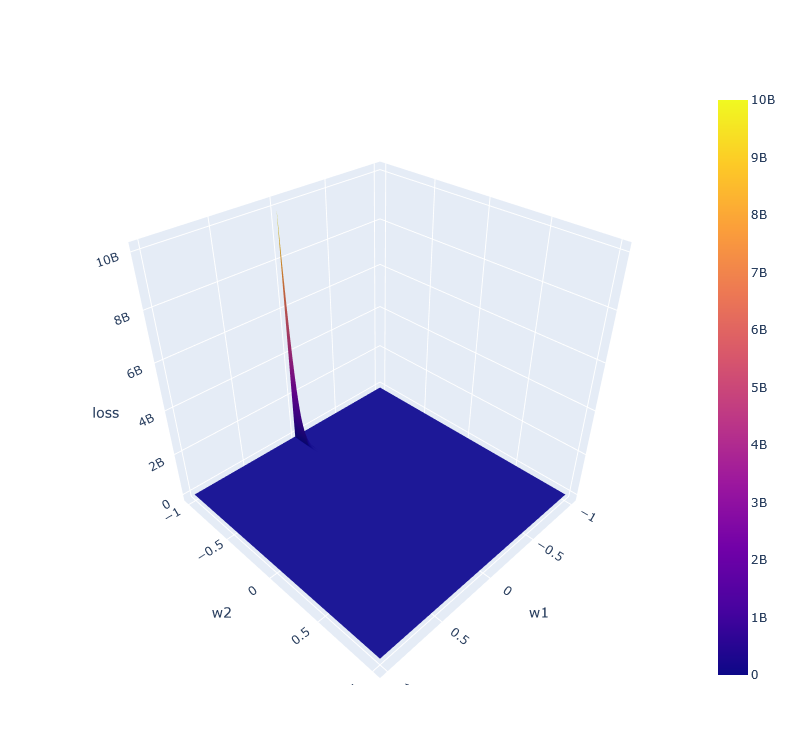}
        \subcaption{NAU-Real NPU (where $\epsilon=1e-5$)}
        \label{fig:loss-landscape-nau-realnpu}
    \end{subfigure}
    \begin{subfigure}{0.8\textwidth}
        \includegraphics[width=\linewidth]{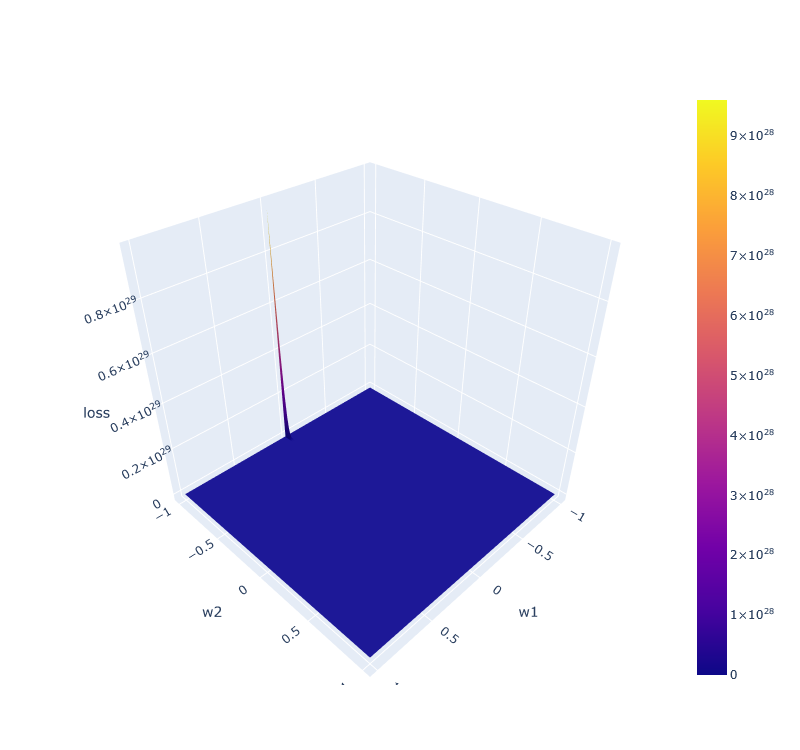}
        \subcaption{NAU-NRU}
        \label{fig:loss-landscape-nau-nru}
    \end{subfigure}
\end{figure}%
\begin{figure}[t]\ContinuedFloat
    \centering
    \begin{subfigure}{0.8\textwidth}
        \includegraphics[width=\linewidth]{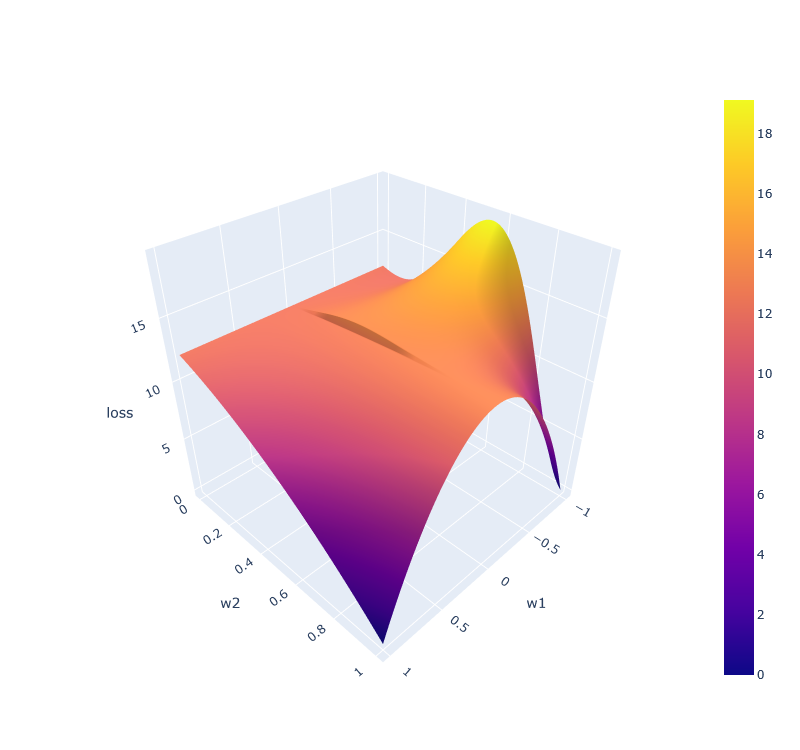}
        \subcaption{NAU-NMRU}
        \label{fig:loss-landscape-nau-nmru}
    \end{subfigure}
\caption{Enlarged loss landscapes of different stacked summative-multiplicative units.}
\label{fig:loss-landscapes}
\end{figure}


\end{document}